\documentclass{SciPost}

\hypersetup{
    colorlinks,
    linkcolor={red!50!black},
    citecolor={blue!50!black},
    urlcolor={blue!80!black}
}

\usepackage{graphicx,amsfonts,amssymb,amsmath,mathrsfs,bm}
\usepackage{algorithm}
\usepackage[noend]{algpseudocode}
\usepackage{float}
\usepackage{listings}
\usepackage{xcolor}

\usepackage[bitstream-charter]{mathdesign}
\DeclareSymbolFont{usualmathcal}{OMS}{cmsy}{m}{n}
\DeclareSymbolFontAlphabet{\mathcal}{usualmathcal}

\fancypagestyle{SPstyle}{
\fancyhf{}
\lhead{\colorbox{scipostblue}{\bf \color{white} ~SciPost Physics Lecture Notes }}
\rhead{{\bf \color{scipostdeepblue} ~Submission }}

\fancyfoot[C]{\textbf{\thepage}}
}

\begin{document}

\pagestyle{SPstyle}

\begin{center}{\Large \textbf{\color{scipostdeepblue}{
Tensor network compressibility of convolutional models
}}}\end{center}

\begin{center}\textbf{
Sukhbinder Singh\textsuperscript{1},
Saeed S. Jahromi\textsuperscript{2,3,4$\star$} and
Rom\'an Or\'us\textsuperscript{2,3,5$\dagger$}
}\end{center}

\begin{center}
{\bf 1} Multiverse Computing, Spadina Ave., Toronto, ON M5T 2C2, Canada
\\
{\bf 2} Donostia International Physics Center, Paseo Manuel de Lardizabal 4, E-20018 San Sebasti\'an, Spain
\\
{\bf 3}Multiverse Computing, Paseo de Miram\'on 170, E-20014 San Sebasti\'an, Spain
\\
{\bf 4} Department of Physics, Institute for Advanced Studies in Basic Sciences (IASBS), Zanjan 45137-66731, Iran
\\
{\bf 5} Ikerbasque Foundation for Science, Maria Diaz de Haro 3, E-48013 Bilbao, Spain
\\[\baselineskip]
$\star$ \href{mailto:email1}{\small saeed.jahromi@iasbs.ac.ir}\,,\quad
$\dagger$ \href{mailto:email2}{\small roman.orus@dipc.org}
\end{center}

\section*{\color{scipostdeepblue}{Abstract}}
\textbf{\boldmath{%
Convolutional neural networks (CNNs) are one of the most widely used neural network architectures, showcasing state-of-the-art performance in computer vision tasks. Although larger CNNs generally exhibit higher accuracy, their size can be effectively reduced by ``tensorization'' while maintaining accuracy, namely, replacing the convolution kernels with compact decompositions such as Tucker, Canonical Polyadic decompositions, or quantum-inspired decompositions such as matrix product states, and directly training the factors in the decompositions to bias the learning towards low-rank decompositions. But why doesn't tensorization seem to impact the accuracy adversely? We explore this by assessing how \textit{truncating} the convolution kernels of \textit{dense} (untensorized) CNNs impact their accuracy. Specifically, we truncated the kernels of (i) a vanilla four-layer CNN and (ii) ResNet-50 pre-trained for image classification on CIFAR-10 and CIFAR-100 datasets. We found that kernels (especially those inside deeper layers) could often be truncated along several cuts resulting in significant loss in kernel norm but not in classification accuracy. This suggests that such ``correlation compression'' (underlying tensorization) is an intrinsic feature of how information is encoded in dense CNNs. We also found that aggressively truncated models could often recover the pre-truncation accuracy after only a few epochs of re-training, suggesting that compressing the internal correlations of convolution layers does not often transport the model to a worse minimum. Our results can be applied to tensorize and compress CNN models more effectively.
}}

\vspace{\baselineskip}

\noindent\textcolor{white!90!black}{%
\fbox{\parbox{0.975\linewidth}{%
\textcolor{white!40!black}{\begin{tabular}{lr}%
  \begin{minipage}{0.6\textwidth}%
    {\small Copyright attribution to authors. \newline
    This work is a submission to SciPost Physics Lecture Notes. \newline
    License information to appear upon publication. \newline
    Publication information to appear upon publication.}
  \end{minipage} & \begin{minipage}{0.4\textwidth}
    {\small Received Date \newline Accepted Date \newline Published Date}%
  \end{minipage}
\end{tabular}}
}}
}


\vspace{10pt}
\noindent\rule{\textwidth}{1pt}
\tableofcontents
\noindent\rule{\textwidth}{1pt}
\vspace{10pt}


\section{Introduction}
\label{sec:intro}
Convolutional Neural Networks (CNNs) are a class of artificial neural networks that excel at computer vision tasks such as image classification and object recognition. The first well-known example of a CNN was LeNet, developed to recognize hand-written digits \cite{lecun1989}. Since then, several influential CNNs have been developed, such as AlexNet\cite{AlexNet}, VGG\cite{VGG}, Inception models\cite{Inceptionv1, Inceptionv3, Inceptionv4}, Xception \cite{Xception}, ResNet\cite{ResNet}, MobileNet\cite{MobileNet} and EfficientNets\cite{EfficientNet}, which have obtained impressive accuracy on extremely large and complex general-purpose image datasets, such as ImageNet\cite{ImageNet}, containing millions of images. CNNs have also been successfully applied to specialized datasets in real-world industrial applications such as defect detection in the manufacturing sector \cite{yang2019real, tabernik2020segmentation, yang2020, bhatt2021image}. 

Like generic neural networks, it is generally expected that the accuracy of CNNs increases as these models become larger \cite{EfficientNet}. This fact is evident in the evolution of state-of-the-art CNN architectures. Modern CNNs, such as ResNet and Inception models, have millions of parameters, and larger models are continuously being developed. On the other hand, larger models are harder to train and deploy on memory-intensive platforms such as mobile phones and embedded systems, e.g., inside autonomous vehicles and robotics. Furthermore, large CNNs (and large NNs in general) are known to be prone to over-fitting. One way to tackle the growing scale of CNNs is to develop sophisticated compression techniques that reduce the number of parameters in a CNN without significantly sacrificing performance.

Several accurate compression techniques have been developed, for example, \textit{pruning, quantization}, and \textit{distillation}. \textit{Pruning} \cite{li2016pruning, he2017channel} removes the weights or filters of the network that have small values and thus have little contribution to the information stored in the network. In \textit{quantization} \cite{Quantization}, the network architecture is left intact, but the numerical precision of the weights is reduced; for instance, double floating point precision may be reduced to single-floating precision or, more drastically, to 8-bit integers. \textit{Knowledge distillation} \cite{DistillationHinton} is a compression method where a larger (teacher) network is used to train and distill its knowledge into a smaller (student) network. 

The main focus of the present paper is a more recent compression technique known as \textit{tensorization}. A neural network is tensorized by replacing the weight matrices inside fully connected layers and/or convolution kernels inside convolution layers with compact tensor decompositions. The most common tensor network decomposition employed for tensorizing a fully connected layer is a matrix product operator (also called tensor train). In contrast, the most common choices of tensorizing a convolution kernel are the Tucker decomposition (including Higher-Order Singular Value Decomposition (HOSVD)) \cite{Tucker1966} and the Canonical Polyadic (CP, CANDECOMP3 or PARAFAC) decomposition \cite{Hars1970}. A general review of these decompositions can be found in Ref.~\cite{TuckerCP}. The effectiveness of Tucker-based \cite{tcnnTucker1, tcnnTucker2, tcnnTucker3} and CP-based \cite{tcnnCP1, tcnnCP2, tcnnCPStable} tensorization of convolution kernels has been demonstrated for several computer vision tasks. While CP and Tucker decompositions are more common tensorizations of convolution layers, quantum-inspired matrix product decompositions have also been applied successfully for several image-related tasks \cite{tcnnTT, tcnnTTandTR, tcnnTR1}. All these decompositions can be described in a unified way as \textit{tensor networks} \cite{Orus2014, Verstraete2008}. 

While traditional compression schemes such as pruning, quantization, and distillation effectively reduce the number of neurons in the network, tensorization compresses the \textit{correlations} between the weights while keeping the number of neurons fixed. (However, tensorization is compatible with these other techniques and can be applied together.) Tensorization adds an \textit{inductive bias} to the learning algorithm towards learning compact, low-rank dense layers or convolution kernels. For instance, successful CNN architectures such as MobileNet and Xception employ depth-wise separable convolutions. This demonstrates that incorporating an explicit bias toward structured convolutions \cite{StructuredConvolutions} can be advantageous rather than restrictive. In this case, a structured (depth-wise separable) convolution can reproduce the accuracy of unstructured convolutions with significantly fewer parameters.

But why does tensorization work at all? A tensorized CNN is essentially different from a dense CNN since replacing dense convolution layers with tensor network decompositions modifies both the loss landscape and the gradient computation during backpropagation. Can we then attribute the success of tensorization to these complex differences? 

In this paper, we provide some evidence that the empirical success of tensorization instead can be accounted for by the structure of correlations between the \textit{trained} weights inside a neural network's layer. We trained two \textit{dense} (untensorized) CNNs -- (i) a small vanilla CNN and (ii) ResNet-50 --- for image classification on CIFAR-10 and CIFAR-100 datasets and then truncated the convolution kernels across various cuts to assess the impact on the accuracy of the model. In both cases, we found that: (1) While the training was not explicitly biased towards low-rank tensors, the convolution kernels can often be compressed along several cuts without significant loss of accuracy, suggesting that this type of \textit{correlation compression} is an \textit{intrinsic} feature of how information is the trained parameters stored in a CNN, not necessarily imposed upon the CNN by introducing an explicit bias during the training. (2)  Compression generally has a more severe impact when applied at initial convolution layers, and (3) Compressed models can often be re-trained for only a few epochs to recover the pre-compression accuracy, suggesting that compressing the internal correlations of convolution layers either largely preserves important information or else translates the model along favorable directions without transporting the model to a worse minimum.

The plan for the rest of the paper is as follows. In Sec.~\ref{sec:background}, we review dense and tensorized CNNs using the picturesque language of tensor networks. We include a brief overview of tensor networks and their graphical calculus in Appendix~\ref{app:A}. Amongst other benefits, the tensor network language provides an intuitive way to estimate the compression of memory required to store a tensorized neural network in memory and the speed up in inference time, both main benefits of tensorization. This cost analysis is described in Appendix \ref{sec:cost}. In Sec.~\ref{sec:truncation}, we describe the correlation truncation of dense CNNs. The results of our truncation experiments for ResNet-50 are presented in Sec.~\ref{sec:resnet}. The results for a vanilla four-layer CNN are postponed to Appendix \ref{sec:vanilla}. We collect our conclusions in Sec.~\ref{sec:conclude}. 

\section{Background} \label{sec:background}

\subsection{Dense Convolutional Neural Networks}
In this section, we briefly review the architecture of CNNs in the context of image classification.

\begin{figure}
    \centering
    \includegraphics[width=0.8\columnwidth]{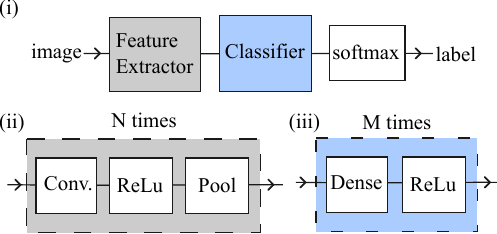}
    \caption{(i) The schematic of a CNN architecture, composed of a feature extractor, classifier, and a Softmax non-linearity that converts the classifier's output into probabilities. (ii) Repeated basic block of operations ($N$ times) that compose the feature extractor. (iii) The basic block of operations repeated ($M$ times) to compose the classifier. While ReLu is a common choice of non-linearity in CNNs, other non-linear functions such as hyperbolic tangent and sigmoid are also used.}
    \label{fig:CNN}
\end{figure}

A CNN has two components, namely, a \textit{feature extractor} and a \textit{classifier}, see Fig.~\ref{fig:CNN}(i).  The feature extractor, Fig.~\ref{fig:CNN}(ii), is a stack of convolutional layers that maps the input image to a set of transformed images (also called feature images). Each convolution layer performs \textit{spatially local} transformations --- \textit{convolutions} and \textit{pooling} --- to learn a set of relevant image features to accomplish the classification task at hand. The features could be simple, such as the location of edges in the image, but in practice, a CNN can learn subtle and complex features of the image. This capability is arguably the main reason underlying the acclaimed success of CNNs in image classification tasks. Each convolutional layer comprises a convolution operation, a pointwise non-linear transformation (such as rectified linear unit, hyperbolic tangent, and sigmoid), and a pooling layer. Transforming an image by a convolution corresponds to passing it through a set of filters, each of which acts on local patches of pixels of the images but is capable of isolating non-local features of the image, see e.g. \cite{UnderstandConv, ConvGuide}. While it is straightforward to define (local) filters that isolate simple global features, such as the location of edges, the filters inside a CNN are parameters learned during the training (or learning) process. On the other hand, pooling coarse-grains an image by applying a simple pixel-reducing function (e.g., maximum or average) to local patches of the image. The features output from the feature extractor are fed to the classifier, a stack of dense, fully connected layers (each fully connected layer is a composition of a linear and a pointwise non-linear, such as ReLu, transformation). The classifier's output is input to a final softmax layer that converts the classifier's output to probabilities over the output labels.\footnote{The softmax layer is not essential in shallow CNNs.} The image label with the largest probability is output by the CNN.

Next, we recall the definitions of convolution and pooling operations for two-dimensional images and describe these fundamental operations as tensor network contractions, which allows us to easily estimate the computational cost of these operations during training and inference. (Higher-dimensional convolutions and poolings can be defined analogously.) 

\subsubsection{Convolution as a tensor network contraction}
In this section, we review convolutions using the language and graphical notation of tensors. Basic concepts and graphical calculus pertaining to tensors and tensor networks are reviewed in Appendix \ref{app:A}. For a more comprehensive review of tensor networks, see, e.g., \cite{Orus2014}.

We begin by reviewing convolutions as they are commonly described. Consider a pixelated image of height $H$ pixels and width $W$ pixels, which we encode into a $H \times W$ real matrix such that each matrix entry equals the intensity of the corresponding pixel. A 2D convolution on the image is a linear transformation of the image matrix. It is described by means of an $X \times Y$ real-valued matrix called a \textit{convolution kernel} (or simply a kernel), and pairs of natural numbers $S \equiv (S^x, S^y)$, and $P \equiv (P^x, P^y)$ called \textit{strides} and \textit{paddings} respectively. Beginning at the image's top-left corner, the kernel matrix is swept across the width and height of the image (matrix), progressing through the sweep in steps equal to the horizontal stride $S^x$ and vertical stride $S^y$. At each step in the sweep, the kernel is convolved with an equal patch of the underlying image. The kernel matrix is first pointwise multiplied (Hadamard multiplication) with an $X \times Y$ patch of the image. The entries of the resulting matrix are then added together, and the sum is stored at a designated pixel location, which is determined by the strides, padding, and kernel dimensions.

If the kernel is not simply a number (i.e., a $1 \times 1$ convolution), the convolution output is usually an image smaller than the input. Therefore, repeated convolutions through a deep CNN can result in very small output images. The usual way to fix this issue is to inflate the input images by padding additional rows and columns around each image matrix. (A common choice is zero padding, corresponding to inserting rows and columns of 0 around the input images.) The number of rows and columns to pad is specified by horizontal and vertical padding sizes, $P^x$ and $P^y$, respectively. 

For example, consider a $6 \times 6$ image $I$,
\begin{equation}
I \equiv \begin{bmatrix}
3 & 0 & 1 & 2 & 7 & 4\\
1 & 5 & 8 & 9 & 3 & 1\\
2 & 7 & 2 & 5 & 1 & 3\\
0 & 1 & 3 & 1 & 7 & 8\\
4 & 2 & 1 & 6 & 2 & 8\\
2 & 4 & 5 & 2 & 3 & 9
\end{bmatrix}	\nonumber
\end{equation}
and a $3 \times 3$ kernel $K$ that detects vertical edges,
\begin{equation}
K \equiv \begin{bmatrix}
1 & 0 & -1\\
1 & 0 & -1\\
1 & 0 & -1
\end{bmatrix}	\nonumber
\end{equation}
Convolving image $I$ with kernel $K$ produces a $4 \times 4$ image $O$,
\begin{equation}
O \equiv \begin{bmatrix}
-5 & -4 & 0 & 8\\
-10 & -2 & 2 & 3\\
0 & -2 & -4 & -7\\
-3 & -2 & -3 & -16
\end{bmatrix}	\nonumber
\end{equation}

The image input to a convolutional layer may contain several channels. For example, a greyscale image consists of only a single (color) channel and an RGB image contains three channels, corresponding to 3 differently colored versions of the images---red, green, and blue, respectively. In each channel (or color in this example), the pixel value quantifies the intensity of the corresponding color at that location in the image. In practice, several kernels may be convolved with a single image in a given channel to produce multiple feature images per image. The total number of images obtained per image is the number of \textit{output channels} that feed into the next convolution layer. 

Next, we describe how convolutions, described above, can be understood as tensor contractions. The input image and the convolution kernel can be expressed as tensors. The input image (and each intermediate feature image) is a 3-index tensor 
\begin{equation}
I_{hw},
\end{equation}
where indices $h={0,1,\ldots, H-1}$ and $w={0,1,\ldots, W-1}$ enumerate pixels along the height and width of the image respectively, and index $={0,1,..., C^{\mbox{\tiny in}}}$ enumerates the input channels. The kernel of a 2D convolution is a 4-index tensor 
\begin{equation}\label{eq:kerneldef}
K_{xy},
\end{equation}
where indices $x=0,1,\ldots, X-1$ and $y=0,1,\ldots,Y-1$ enumerate the rows and columns of each convolution kernel (that is, for fixed input and output channels). We remark that a set of multiple kernels, as $K$ here, is sometimes called a \textit{filter}. But in this paper, we use ``filter'' and ``kernel'' interchangeably.

\begin{figure}
    \centering
    \includegraphics[width=0.8\columnwidth]{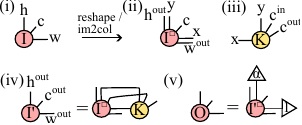}
    \caption{(i) The input image as a 3-index tensor $I$. (ii) The ``patch image'' tensor $I^\square$ obtained from the input image either by reshaping indices (for simple convolutions) or applying a more general shuffling transformation called \textbf{im2col}() \cite{Im2col, Im2colMatLab}.
    (iii) The convolution kernel $K$ as a 4-index tensor.
    (iv) Convolution on an image as a contraction of the patch image tensor $I^\square$ with the convolution kernel tensor $K$, producing an output feature image $I'$ \cite{ConvTN}. A non-linearity such as ReLu (not shown here) is applied on $I'$ to obtain a transformed image. (v) The resulting feature image $I'$ is reorganized into a patch image $I'{}^\square$. Then average pooling can be understood as a tensor contraction of $I'{}^\square$ with a vector whose components are all ones and another vector whose components are all $\alpha$ [see Fig.~\ref{fig:tensors}(v)].}
    \label{fig:convAsTN}
\end{figure}

Having expressed the image and the kernel as tensors, the convolution operation can now be understood as a tensor contraction. To see this, consider the relatively simple convolution corresponding to the following choices (*): 
\begin{enumerate}
    \item Image is square, $H = W$,
    \item Kernel is square, $X = Y$,
    \item $H$ is divisible by $X$, and $S^x = S^y = X$ (the stride in both directions is equal to $X$)
    \item No padding.
\end{enumerate}
Such a convolution can be executed by dividing the image into a $\frac{H}{X} \times \frac{H}{X}$ grid of non-overlapping $X \times X$ patches and then taking the elementwise product of each patch with the kernel matrix, summing all the entries of the resulting matrix and assigning it to the corresponding entry of the output image matrix, which has dimensions $\frac{H}{X} \times \frac{H}{X}$.

The above convolution, corresponding to the choices (*), can be described as a tensor contraction. First, we reshape\footnote{We use reshape to mean the operation implemented by the eponymous NumPy function.} the input image $I$ into an intermediate 5-index \textit{patch tensor} $I^{\square}$ with dimensions $\frac{H}{X} \times X \times \frac{H}{X} \times X \times C^{\mbox{\tiny in}}$, see Fig.~\ref{fig:convAsTN}(ii). Thus, the input image is reshaped into a stack of image patches, each of which will be convolved with the kernel. The two steps in the convolution -- the elementwise multiplication between the image pixels and the kernel matrix and subsequent sum of the matrix entries --- can both be expressed as the tensor network contraction. Elementwise (Hadamard) matrix multiplication can be realized as a tensor contraction as depicted in Fig.~\ref{fig:tensors}(ix). The summation of the resulting matrix elements is equivalent to contracting each of matrix index with a vector of ones, as depicted in Fig.~\ref{fig:tensors}(xii). The copy tensors introduced by the Hadamard multiplication can be eliminated using the equalities depicted in Fig.~\ref{fig:tensors}(vii). Thus, we arrive at the tensor network contraction depicted in Fig.~\ref{fig:convAsTN}(iv) that implements the convolution for the choices (*).

More general convolutions, corresponding to choosing parameters differently than (*), can also be understood as tensor contractions. In this case, the 5-index patch tensor $I^{\square}$ is not obtained by a simple reshape of $I$ -- in fact, tensor $I^{\square}$ is generally bigger than $I$ --, instead, the various patches are constructed by extracting and re-arranging elements of the image $I$ (which may be padded). This operation can be implemented by brute force by scanning all the elements of $I$ and allocating them to designated patches. However, the construction of the patch tensor can also be carried by a fast and efficient operation known as \textbf{im2col} \cite{Im2col}, whose implementation is available in several standard libraries, e.g., \cite{Im2colMatLab}. Once the patch tensor $I^{\square}$ is constructed -- either by brute force or by using \textbf{im2col} --, the convolution is carried out the same contraction shown in Fig.~\ref{fig:convAsTN}(iv). The contraction results in an output image 
\begin{equation}
I'_{h^{\mbox{\tiny out}}w^{\mbox{\tiny out}}},
\end{equation}
of dimensions $H^{\mbox{\tiny out}} \times W^{\mbox{\tiny out}}$. The dimensions of the output image are determined by the input dimensions, strides, and padding according to the following formulas:
\begin{align}\label{eq:convoutdims}
    H^{\mbox{\tiny out}} &= \frac{H - X + 2P^x}{S^x} + 1 \\
    W^{\mbox{\tiny out}} &= \frac{W - Y + 2P^y}{S^y} + 1
\end{align}
A pointwise non-linear transformation (such as ReLu) is applied to the $I'$ before it is input to a pooling layer.

\subsubsection{Average pooling as a tensor contraction}
Pooling is a coarse-graining procedure to reduce the size of the feature images obtained after convolution. However, a pooling operation is quite similar to a convolution. We slide a window over the image and apply a function such as max or average to the entries of the image inside the window. The result (a real number) is assigned to a designated location in the output, again determined by the window size, strides, and padding, producing an output image of dimensions determined according to Eq.~\ref{eq:convoutdims}.

In particular, average pooling is a convolution with a constant kernel matrix whose entries are equal to $\frac{1}{N}$ where $N$ is the window size. (This constant kernel is equal to the outer product of a vector of all ones times a vector of all $\alpha=\frac{1}{N}$s.) Therefore, average pooling can also be understood as a tensor contraction, just like a convolution. (However, max pooling cannot be expressed as a contraction.) First, we reorganize the image $I'$ output from the convolution layer (which includes the non-linear transformation) into a 5-index patch tensor $I'^{\square}$ using either reshape when the choices (*) apply or, more generally, \textbf{im2col}. The entries of each patch inside the patch tensor $I'^{\square}$ are then averaged. The average of a $N \times N$ matrix can be obtained by contracting one of the matrix indices with a vector of ones and the other with a vector whose entries are all $\alpha = 1/N$ ($N$ is the window size), as depicted in Fig.~\ref{fig:tensors}(xiii). The total contraction that implements the average pooling is shown in Fig.~\ref{fig:convAsTN}(v).

\subsection{Tensor network decompositions of convolution layers}
\label{sec:tensorize-CNN}
Having described convolution kernels as tensors and the central operations of convolution and pooling as tensor contraction, we now review tensorization of convolution layers. A convolution layer is tensorized by replacing the convolution kernel with a low-rank tensor decomposition, which is then exploited to accelerate convolution and pooling operations. 

Popular choices of tensor decompositions are the Tucker and CP decompositions. These are reviewed below for the case of 2D convolutions, but the decompositions are readily generalized to higher-order convolutions. We also briefly review ``mode-mixing'' tensor network decompositions -- such as MPS/tensor train and tensor ring -- that have been shown to be effective for some image classification tasks.

\subsubsection{Tucker Decomposition}
\begin{figure}
    \centering
    \includegraphics[width=8cm]{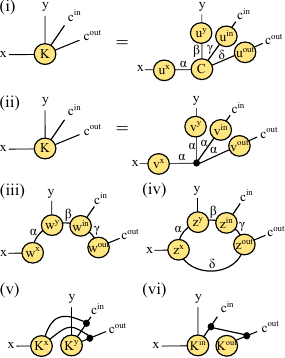}
    \caption{Popular tensor network decompositions of the convolution kernel. (i) Tucker decomposition. (ii) CP decomposition. Matrix Product State-based decompositions: (iii) Tensor Train (MPS with open boundary condition) and (iv) Tensor Ring (MPS with periodic boundary condition). Examples of structured convolutions (Adapted from \cite{StructuredConvolutions}): (v) a convolution kernel comprised of rank-1 filters, namely, spatially separable convolution), and (vi) a depthwise separable convolution kernel underlying successful CNNs such as Xception \cite{Xception} and MobileNet \cite{MobileNet}.}
    \label{fig:tensordecompositions}
\end{figure}

The Tucker decomposition of the 4-index kernel $K_{xy}$ for a 2D convolution is given by:
\begin{equation}\label{eq:Tucker}
    K_{xy} = \sum_{\alpha\beta\gamma\delta}C_{\alpha\beta\gamma\delta} U^{X}_{x\alpha} U^{Y}_{y\beta} U^{\mbox{\tiny in}}_{\gamma} U^{\mbox{\tiny out}}_{\delta}.
\end{equation}
Here, $K$ is decomposed into a 4-index core tensor $C$, and four \textit{mode matrices}, $U^{X}, U^{Y}, U^{\mbox{\tiny in}},$ and $U^{\mbox{\tiny out}}$. See Fig.~\ref{fig:tensordecompositions}(i). The dimensions $|\alpha|, |\beta|, |\gamma|, |\delta|$ of the internal indices of the decomposition are called the \textit{Tucker ranks}.

The effectiveness of Tucker decomposition-based tensorization of CNNs has been demonstrated in Refs.~\cite{tcnnTucker1, tcnnTucker2, tcnnTucker3}. For instance, Ref.~\cite{tcnnTucker1} achieved lossless accuracy with a compression ratio\footnote{The compression ratio is the ratio of the size of the dense and tensorized models.} of 5.46$\times$ on AlexNet, 7.40$\times$ on VGG-S, and 1.09$\times$ on VGG-16, and Ref.~\cite{tcnnTucker3} achieved lossless accuracy for ResNet-18 on CIFAR-10 with a CR and speedup of 11.82$\times$ and 5.48$\times$, respectively.

\subsubsection{Higher-order singular value decomposition}
\textit{Higher-Order Singular Value Decomposition} (HOSVD) \cite{delathauwer2000, DeLathauwer2006} is a specific instance of Tucker decomposition. A HOSVD of a tensor $T$ is a Tucker decomposition in which the mode matrices and core tensor fulfill certain orthogonality constraints (explained below). In this section, we first describe the HOSVD of the convolution kernel and then describe how the orthogonality constraints can be preserved when training an HOSVD-based tensorized CNN.

\begin{figure}
    \centering
    \includegraphics[width=8cm]{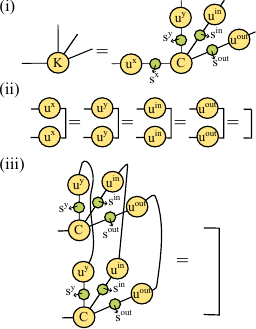}
    \caption{(i) The HOSVD decomposition of the convolution kernel. $S^X, S^Y, S^{\mbox{\tiny in}}$ and $S^{\mbox{\tiny out}}$ are diagonal matrices whose diagonal entries are the singular values. We also refer to these as the \textit{single-mode} singular values of the kernel, corresponding to the four modes (indices) of the kernel.  (ii) The mode matrices $U^X, U^Y, U^{\mbox{\tiny in}}$, and $U^{\mbox{\tiny out}}$ are orthogonal, fulfilling, $U^X (U^X)^T = I_{|\alpha|}$, and so on. Here, ${(.)}^T$ denotes matrix transposition. The size of the identity matrix depicted on the right is equal to the HOSVD rank of that mode. (iii) The part of the HOSVD obtained by discarding any orthogonal mode matrix is an isometry; namely, it fulfills an identity similar to the one shown in this panel for mode $x$.}
    \label{fig:hosvd}
\end{figure}

The HOSVD decomposition of the kernel is (see Fig.~\ref{fig:hosvd}(i)):
\begin{equation}\label{eq:hosvd}
\begin{split}
    K_{xy} = \sum_{\alpha\beta\gamma\delta}& C_{\alpha\beta\gamma\delta} (U^{X}_{x\alpha} S^{X}_{\alpha\alpha})~ (U^{Y}_{y\beta} S^{Y}_{\beta\beta}) \\
    &(U^{\mbox{\tiny in}}_{\gamma} S^{\mbox{\tiny in}}_{\gamma\gamma})~ (U^{\mbox{\tiny out}}_{\delta} S^{\mbox{\tiny out}}_{\delta\delta}).
\end{split}
\end{equation}
where $S^X, S^Y, S^{\mbox{\tiny in}}, S^{\mbox{\tiny out}}$ are diagonal matrices with non-negative diagonal entries, and the mode matrices are orthogonal, namely, 
\begin{equation}\label{eq:modeconstraints}
\begin{split}
&U^X (U^X)^T = I_{|\alpha|}, U^Y (U^Y)^T = I_{|\beta|}, \\
&U^{\mbox{\tiny in}} (U^{\mbox{\tiny in}})^T = I_{|\gamma|},~U^{\mbox{\tiny out}} (U^{\mbox{\tiny out}})^T = I_{|\delta|}.
\end{split}
\end{equation}
Here, $(.)^T$ denotes matrix transposition, and $I_n$ denotes the $n \times n$ identity matrix. These mode constraints are depicted in Fig.~\ref{fig:hosvd}(ii). Note that our definition of HOSVD, Eq.~\ref{eq:hosvd}, differs slightly from the standard definition, in which the diagonal matrices $S$s are multiplied into the core tensor $C$. 

\subsubsection{Canonical Polyadic Decomposition}
The CP decomposition of an $n$-index tensor is a sum of tensor products of $n$ vectors, generalizing the structure of eigenvalue decomposition of matrices to higher-dimensional tensors. The CP decomposition of the kernel $K$ of a 2D convolution takes the form:
\begin{equation}\label{eq:cp}
    K_{xy} = \sum_{\alpha=1}^{r} V^{x}_{x\alpha} V^{y}_{y\alpha} V^{\mbox{\tiny in}}_{\alpha} V^{\mbox{\tiny out}}_{\alpha},
\end{equation}
where $r$ is called the CP rank. The CP decomposition is a constrained instance of the Tucker decomposition, where the core tensor $C$ is constrained to be the delta tensor, namely,  $C_{\alpha\beta\gamma\delta} = \delta_{\alpha\beta} \delta_{\beta\gamma} \delta_{\gamma\delta}$. 

Successful applications of CP decomposition-based tensorization of CNNs have been demonstrated by Refs.~\cite{tcnnCP1, tcnnCP2, tcnnCPStable}. However, determining an approximate CP decomposition with a given rank $k$ is an NP-hard problem \cite{CPRankIsHard}, which limits the accuracy of CP compression of trained models (prior to fine-tuning). Furthermore, the usual CP decomposition suffers from a known instability issue---fitting the convolutional tensors by numerical optimization algorithms often encounters diverging components \cite{CPIsUnstable, tcnnCP2}. This issue was tackled in Ref.~\cite{tcnnCPStable}, where the authors proposed a stable but sophisticated algorithm to approximate and fine-tune the CP decomposition, demonstrating the validity of their method by compressing VGG-16 (1.10$\times$, 5.26$\times$), ResNet-18 (3.82$\times$, 3.09$\times$), and ResNet-50 (2.51$\times$, 2.64$\times$) models for the ILSVRC-12 dataset. In parentheses, listed is the reduction in the number of weights and the speedup, respectively.

\section{Truncation of dense CNNs} \label{sec:truncation}

This paper aims to assess how truncating convolution layers impacts the accuracy of dense-trained CNNs. Specifically, we consider two ways of truncating the convolution kernel $K$ (Eq.~\ref{eq:kerneldef}) based on (i) SVD (including and beyond HOSVD, Eq.~\ref{eq:hosvd}) and (ii) CP decomposition (Eq.~\ref{eq:cp}). We quantify the truncation by measuring the reduction in the Frobenius norm of the kernel.

\subsection{Single-mode truncations}
To apply the SVD-based truncation, we first choose a bipartition of the indices of the convolution kernel $K_{xy}$. A \textit{single-mode} truncation corresponds to a bipartition between a single mode (index) of $K$ and the remaining three indices. First, $K$ is reshaped into a matrix according to that bipartition of indices; for instance, choosing index $x$ corresponding to the kernel width, we can reshape $K$ into a matrix 
\begin{equation}
M^{\mbox{\tiny KW}}_{x\alpha} = K_{x(y)} \nonumber
\end{equation}
where the indices $y, , $ are group into an combined index $\alpha$. (The subscript KW stands for Kernel Width.) We then perform the SVD $M^{\mbox{\tiny KW}} = USV$ where matrices $U$ and $V$ are isometries, namely, $U^\dagger U = V V^\dagger = I$, and $S$ is a diagonal matrix with non-negative diagonal entries called the singular values of $M^{\mbox{\tiny KW}}$. In terms of components, we have 
\begin{equation}\label{eq:example_trunc}
M^{\mbox{\tiny KW}}_{i\alpha} = \sum_{k=1}^{n} U_{ik}S_{kk}V_{k\alpha},  
\end{equation}
where $n \leq \mbox{min}(|i|, |\alpha|)$ is the total number of singular values. (Here, $|i|$ denotes the size of index $i$, see Appendix \ref{app:A}.) In this instance, $|i|$ is the kernel width, and $|\alpha|$ is the product of kernel height, number of input channels, and number of output channels.
Note that the (square) Frobenius norm of matrix $M^{\mbox{\tiny KW}}$ is the sum of the square of its singular values, $\lVert M^{\mbox{\tiny KW}} \rVert = \sum_{k=1}^{n} = S_{kk}^2$, since 
\begin{align}
\lVert M^{\mbox{\tiny KW}} \rVert &= \mbox{Tr}(M^{\mbox{\tiny KW}}M^{\mbox{\tiny KW}}{}^\dagger) \\\nonumber
&= \mbox{Tr}(USVV^\dagger S U^\dagger) = \mbox{Tr}(S^2). 
\end{align}
In an SVD-based truncation, we discard the lowest $\phi$ number of singular values (in other words, preserve the largest $n - \phi$ number of singular values) and multiply back the truncated SVD to obtain the truncated matrix $\tilde{M}^{\mbox{\tiny KW}, \phi}$. The norm of the truncated matrix $\tilde{M}^{\mbox{\tiny KW}, \phi}$ is, therefore, 
\begin{equation}\label{eq:trunc}
\lVert \tilde{M}^{\mbox{\tiny KW}, \phi} \rVert = \sum_{k=1}^{n-\phi} S_{kk}.
\end{equation}
Finally, the truncated matrix is reshaped to obtain a four-index truncated kernel $\tilde{K}^{\mbox{\tiny KW}, \phi}$.

\subsection{Two-mode truncations}
A \textit{two-mode} SVD-based truncation proceeds similarly after grouping pairs of indices of $K$. For instance, let us consider the bipartition of the indices (modes) of $K$ in which the modes kernel width (KW) and number of input channels (IN) are paired together. We first permute indices of $K$ so that indices $x$ and $y$ become first neighbors and then group them by reshaping:
\begin{equation}
M^{\mbox{\tiny KW, IN}} = K_{(x)(y)}. \nonumber
\end{equation}
We then proceed similar to the single-mode case by singular value decomposing $M^{\mbox{\tiny KW, IN}}$, discarding $\phi$ smallest singular values, multiplying together the truncated SVD to obtain the truncated matrix $\tilde{M}^{\mbox{\tiny KW, IN}, \phi}$, and finally reshaping it into a 4-index tensor and permuting indices to the original order to obtain a truncated kernel $\tilde{K}^{\mbox{\tiny KW, IN}, \phi}$.

\subsection{MPS-based truncation}
We remark that single-mode and two-mode SVD-based truncation taken together also evaluate the impact of applying a tensor train/MPS-based truncation since each internal index of the MPS corresponds to a bipartition of the convolution kernel indices.

\subsection{CP-based truncation}
The CP-based truncation proceeds by replacing convolution kernel $K$ with a truncated kernel $\tilde{K}^{\mbox{\tiny CP}, r}$ that is reconstituted from a CP decomposition (with given rank $r$) of $K$, Eq.~\ref{eq:cp}. The truncation error is controlled by the CP rank $r$ and typically reduces as $r$ increases.

\subsection{Quantifying the impact of truncations}
We define the \textit{norm loss} as the percentage reduction in the norm after truncation (using either SVD or CP decomposition):
\begin{equation}\label{eq:normloss}
\mbox{\it{Norm loss }} \% = \frac{ \lVert K \rVert - \lVert \tilde{K} \rVert}{\lVert K \rVert} \times 100.
\end{equation}
In the SVD-based truncation, we have $\lVert \tilde{K} \rVert = \lVert \tilde{M} \rVert$ where $\tilde{M}$ is the matrix representation of truncated kernel across any index bipartition.

The bipartite correlations between the kernel modes can be quantified using the quantum-inspired measure called \textit{entanglement entropy}. Given a bipartition of the modes (indices) of $K$, and the SVD of the corresponding bipartite matrix $M = USV$ (e.g., $M$ could be $M^{\mbox{\tiny KW}}$ or $M^{\mbox{\tiny KW,IN}}$ introduced above) the entanglement entropy $E(M)$ is given by
\begin{equation}\label{eq:ee}
E(M) = -\mbox{Tr} (\hat{S}~\mbox{ln} \hat{S}),~~~\hat{S} \equiv \frac{S}{\sqrt{\lVert S \rVert}},
\end{equation}
where $\hat{S}$ are the normalized singular values. The entanglement entropy $E(M)$, a positive real number, quantifies correlations between pairs of modes in the convolution kernel (regarded as a quantum state); a larger value of $E(M)$ indicates a larger amount of correlations. Entanglement entropy is zero if and only if there are no correlations between the chosen pairs of the modes, namely, the matrix $M$ decomposes as a tensor product $M = U \otimes V$.

Both SVD and CP decomposition-based truncations can be understood as truncating correlations between the weights --- more specifically, the modes in which the weights are organized --- of the CNN since the internal degrees of freedom exposed by these decompositions carry correlations between the weights. We can quantify the correlations lost after a truncation (SVD or CP-based) by means of the percentage correlation loss,
\begin{equation}\label{eq:corrloss}
\mbox{\it{Corr. loss }} \% = \frac{E(M) - E(\tilde{M})}{E(M)} \times 100.
\end{equation}
However, in practice, the norm loss, Eq.~\ref{eq:normloss}, is often proportional to the correlation loss, Eq.~\ref{eq:corrloss}. This is apparent when considering the SVD-based truncation since discarding singular value directly reduces the norm, Eq.~\ref{eq:trunc}, and the entanglement entropy, Eq.~\ref{eq:ee}. We found this to be the case also for CP-based truncations, which is sensible since the CP-rank of the kernel is an upper bound on the largest single-mode SVD rank \footnote{See Theorem 4.2 on https://www.tensors.net/tutorial-4}. Therefore, we only tracked the norm loss in our numerical experiments.

Finally, we define the \textit{compression ratio} of a kernel truncation as the ratio of the number of parameters required to specify the kernel after and before the truncation, respectively. For example, for an SVD-based truncation for mode KW, the compression ratio is $\frac{\tilde{M}^{\mbox{\tiny KW, IN}, \phi}}{\lVert M^{\mbox{\tiny KW}} \rVert}$.

\section{Results for ResNet-50}\label{sec:resnet}
Next, we report the results of our truncation experiments. With these experiments, we aimed to address questions such as 
\begin{itemize}
\item Are CNNs robust against correlation truncations based on SVD and CP? (This would add to the evidence in favor of tensorizing CNNs.)
\item How does the impact of SVD-based truncation compare across different cuts of the convolution kernel?
\item How does the impact of truncation vary with the depth of the convolution layer in the network?
\item How does CP decomposition-based truncation compare with SVD-based truncation?
\item How quickly does accuracy recover after truncation?
\end{itemize}
We performed two sets of experiments to assess the impact of correlation truncation --- one for ResNet-50 trained on CIFAR-10 and CIFAR-100 datasets and one for a vanilla four-layer CNN. Within these choices, we can assess the truncation impact on two different CNN architectures, ResNet and Vanilla, trained on the same dataset, CIFAR-10, and the same CNN architecture, ResNet-50, trained on two different datasets. Besides being different in size, an important difference between ResNet-50 and our vanilla CNN is that ResNet has residual or skip connections, which could potentially endow some resilience to truncations. We observed that residual connections allowed some kernels to vanish during the fine-tuning training, which increased the impact of truncating nearby layers. However, we found the overall robustness trend was comparable in both cases, undermining any special role of residual connections in these experiments.

ResNet-50 \cite{ResNet} is a 50-layer residual convolutional neural network comprised of 48 convolutional layers, one MaxPool layer, and one average pool layer. The layers are organized into residual bottleneck blocks; each residual block adds to the residue stream of the input fed into the block, while the bottleneck corresponds to the presence of $1 \times 1$ convolutions. We fine-tuned a pre-trained ResNet-50 for the ImageNet dataset for the smaller CIFAR-10 and CIFAR-100 datasets by modifying the initial convolution layer to match the CIFAR image size and the classifier to match the output size to 10 and 100 for CIFAR-10 and CIFAR1-100 respectively.

The CIFAR-10 dataset consists of 60000 $32 \times 32$ color images in 10 classes (\textit{airplanes}, \textit{cars}, \textit{birds}, \textit{cats}, \textit{deer}, \textit{dogs}, \textit{frogs}, \textit{horses}, \textit{ships}, and \textit{trucks}), with 6000 images per class. There are 50000 training images and 10000 test images. The CIFAR-100 dataset is similar to the CIFAR-10 but has 100 classes containing 600 images each. Each class has 500 training images and 100 testing images. The 100 classes (e.g., \textit{apples},  \textit{clock},  \textit{bed}) are grouped into 20 superclasses (e.g.,  \textit{food containers},  \textit{fruit and vegetables},  \textit{household electrical devices}). Each image is labeled with its class and superclass.

\subsection{Setup of the truncation experiments}
\begin{figure}
    \centering
    \includegraphics[width=0.8\columnwidth]{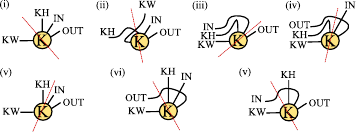}
    \caption{All the bipartitions of the kernel modes (indices) considered in this paper for truncation. Single-mode bipartitions labeled (i) KW, (ii) KH, (iii) OUT, and (iv) IN. Two-mode bipartitions labeled (v) OUT, IN, (vi) OUT, KW, and (vii) OUT, KH. For each picture, the kernel can be transformed into a corresponding matrix by bending, crossing, and grouping indices, as shown.}
    \label{fig:cuts}
\end{figure}

For our truncation experiments, we selected two convolution layers from each bottleneck block inside ResNet-50, a total of eight convolution layers, approximately evenly spaced across the depth of the network. We refer to eight convolution layers as \textit{conv1}, \textit{conv2}, $\ldots$, and \textit{conv8} in increasing order of depth. Each convolution kernel is a four-mode (index) tensor; we denote the four modes by:-
\begin{enumerate}
\item OUT: number of output channels,
\item IN: number of input channels,
\item KW: kernel width,
\item KH: kernel height.
\end{enumerate}
We consider various bipartitions of these four modes. With a slight abuse of notation, we specify a bipartition simply by listing the modes grouped on one side. For instance, the label OUT, IN denotes the bipartition (OUT, IN) -- (KW, KH), where mode pairs OUT and IN have been grouped on one side and the remaining modes, KW and KH, on the other. Similarly, bipartition labeled OUT corresponds to (OUT) -- (IN, KW, KH) where modes IN, KW, and KH have been grouped. Accordingly, we consider the four \textit{single-mode} bipartitions, see Fig.~\ref{fig:cuts}(i)-(iv), in which only a single index/modes appears on one side, namely, OUT, IN, KW, and KH, and three \textit{two-mode} bipartitions, namely, OUT,IN; OUT,KW; and OUT, KH; see Fig.~\ref{fig:cuts}(v)-(vii).

We can reshape the convolution kernel $K$ into a matrix $M$ according to bipartition. For instance, $M^{\mbox{\tiny OUT}}$ denotes the matrix obtained by reshaping $K$ according to the bipartition labeled OUT. The SVD-based truncation proceeds by truncating the singular value spectrum of these matrices.
In our experiments, we found that the results for the SVD-based truncation for the following pairs of bipartitions were very similar:
\begin{enumerate}
\item Bipartition OUT and Bipartition IN
\item Bipartition KW and Bipartition KH
\item Bipartition OUT, KW and Bipartition OUT, KH
\item Bipartition IN, KW and Bipartition IN, KH
\end{enumerate}
Therefore, to avoid repetition, we have omitted results for one bipartition in each of the above pairs. For SVD-based truncation, we chose a range for truncating the number of singular values for each of the eight convolution layers. The ranges were determined according to the size of the convolution kernel, as summarized in the following tables.
\begin{table}[h]
\centering
 \begin{tabular}{||c|c|c|c|c|} 
 \hline
 Kernel size & OUT & IN & KW & KH \\ [0.5ex] 
 \hline\hline
 16,16,3,3 & 2:14:2 & 2:14:2 & 1:2:1 & 1:2:1 \\ 
 32,32,3,3 & 3:30:3 & 3:30:3 & 1:2:1 & 1:2:1 \\
 64,64,3,3 & 10:60:10 & 10:60:10 & 1:2:1 & 1:2:1 \\
 128,128,3,3 & 20:120:20 & 20:120:20 & 1:2:1 & 1:2:1 \\
 256,256,3,3 & 50:250:50 & 50:250:50 & 1:2:1 & 1:2:1 \\
 512,512,3,3 & 50:500:50 & 50:500:50 & 1:2:1 & 1:2:1 \\ [1ex] 
 \hline
 \end{tabular}
 \caption{Range of truncations --- the number of largest singular values to keep; equal to $n-\phi$ in Eq.~\ref{eq:trunc} --- for single-mode bipartitions. Each range is specified as start~:~end~:~step size.}
\end{table}

\begin{table}[h]
\centering
 \begin{tabular}{||c|c|c|c|} 
 \hline
 Kernel size & OUT-IN & OUT-KW & OUT-KH \\ [0.5ex] 
 \hline\hline
 16,16,3,3 & 1:8 & 2:14:2 & 1:2:1 \\ 
 32,32,3,3 & 1:8 & 3:30:3 & 1:2:1 \\
 64,64,3,3 & 10:60:10 & 10:60:10 & 1:2:1 \\
 128,128,3,3 & 20:120:20 & 20:120:20 & 1:2:1 \\
 256,256,3,3 & 50:250:50 & 50:250:50 & 1:2:1 \\
 512,512,3,3 & 50:500:50 & 50:500:50 & 1:2:1 \\ [1ex] 
 \hline
 \end{tabular}
 \caption{Range of truncations for two-mode bipartitions. Each range is specified as start:~end:~step size.}
\end{table}
For CP decomposition-based truncation, we chose CP ranks within the range 10:40:10 for \textit{conv1, ..., conv4} and 10:30:10 for \textit{conv5, ..., conv8}. When computing CP decompositions for ranks larger than these values, we encountered numerical instabilities.

\subsection{Spectra of the convolution kernels}

\begin{figure}
    \centering
    \includegraphics[width=\columnwidth]{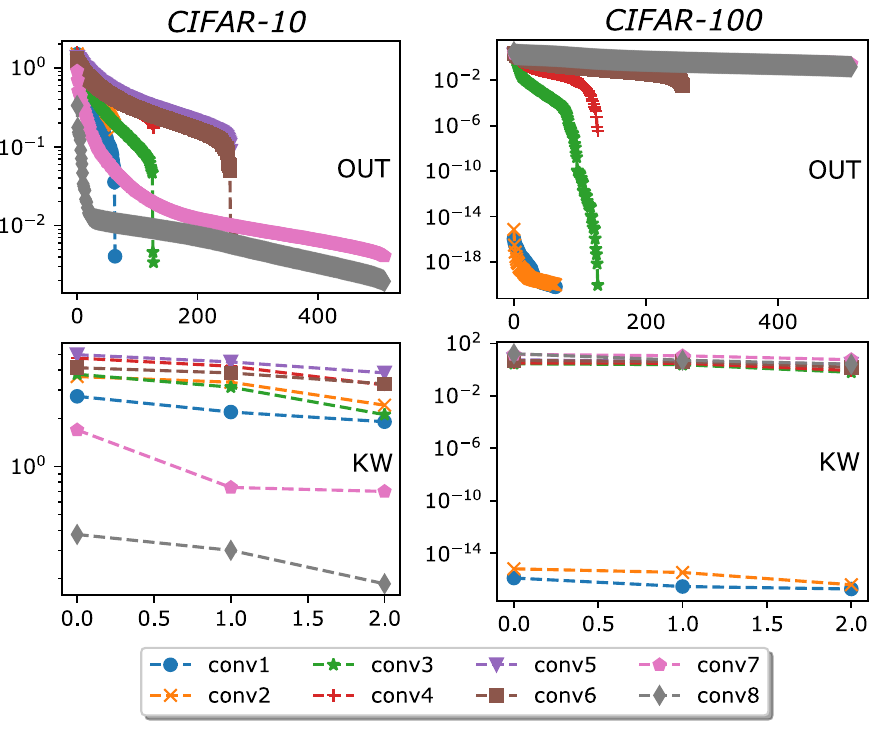}
    \caption{Singular values of the convolution kernels for bipartitions OUT (top) and KW (bottom). The spectra are quite flat and do not contain many small singular values, implying that our trained dense CNNs do not have low ranks, which could otherwise account for the robustness against truncations. Note that for CIFAR-100, \textit{conv1} and \textit{conv2} have nearly vanishing norms, enabled by residual/skip connections in the network, allowing information to bypass these layers. In this instance, \textit{conv3} should be understood as effectively the shallowest layer of the model and the impact of truncating it should be calibrated as such.}
    \label{fig:single-mode-spec}
\end{figure}

First, we plot the singular value spectra of the convolution kernels across various bipartitions in Fig.~\ref{fig:single-mode-spec}. The plot in the top-left panel shows the spectrum corresponding to the bipartition OUT, namely, the singular values of the kernel reshaped as a matrix by combining the modes IN, KW, and KH. The top-right, bottom-left, and bottom-right panels show the singular values for the bipartitions IN, KW, and KH, respectively. For these bipartitions, we see that the spectra are quite flat and do not contain many small singular values, implying that our trained dense CNNs do not have low singular value ranks across these bipartitions to begin with, which could otherwise account for the robustness against correlation truncation across these bipartitions. Notice that for CIFAR-100, we found that the trained \textit{conv1} and \textit{conv2} kernels have almost vanishing norms, enabled by the presence of residual or skip connections in ResNet-50, which allow the information flow to bypass these convolution layers.

\begin{figure}
    \centering
    \includegraphics[width=7cm]{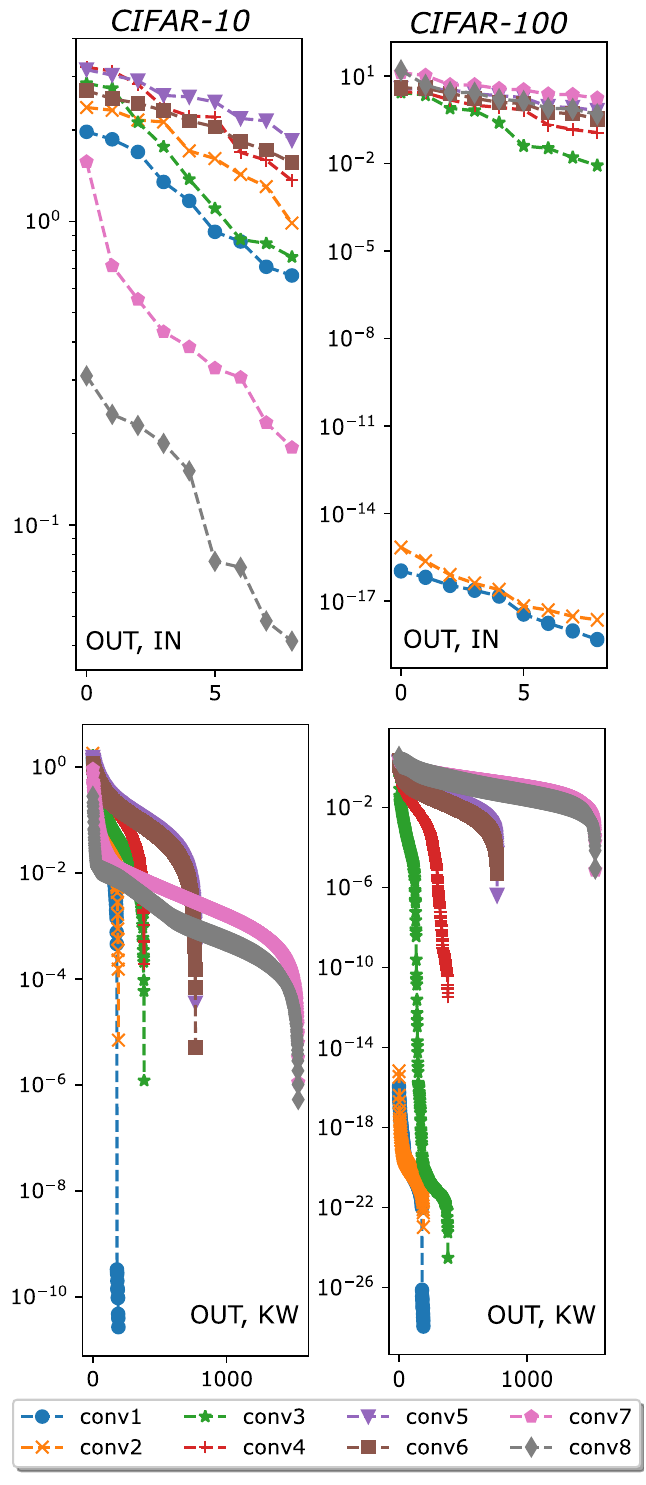}
    \caption{Singular values of the convolution kernels for bipartitions OUT, IN (top) and OUT, KW (bottom). The spectra are quite flat and do not contain many small singular values, implying that our trained dense CNNs do not have low ranks, which could otherwise account for the robustness against truncations. Note that for CIFAR-100, \textit{conv1} and \textit{conv2} have nearly vanishing norms, enabled by residual/skip connections in the network, allowing information to bypass these layers.}
    \label{fig:pairwise_spectrum_combined}
\end{figure}

The top panels in Fig.~\ref{fig:pairwise_spectrum_combined} show the spectrum across the bipartition OUT-IN. The spectrum is once again quite flat and not compressible without norm loss. However, the spectrum for bipartition OUT, KW contains several small singular values that can be discarded without significant norm loss. Accordingly, we chose sufficiently broad truncation ranges for this bipartition (and the bipartition OUT, KW, which exhibits similar spectra) to effect a significant norm loss.

\subsection{Single-layer correlation truncation}

\begin{figure}
    \centering
    \includegraphics[width=\columnwidth]{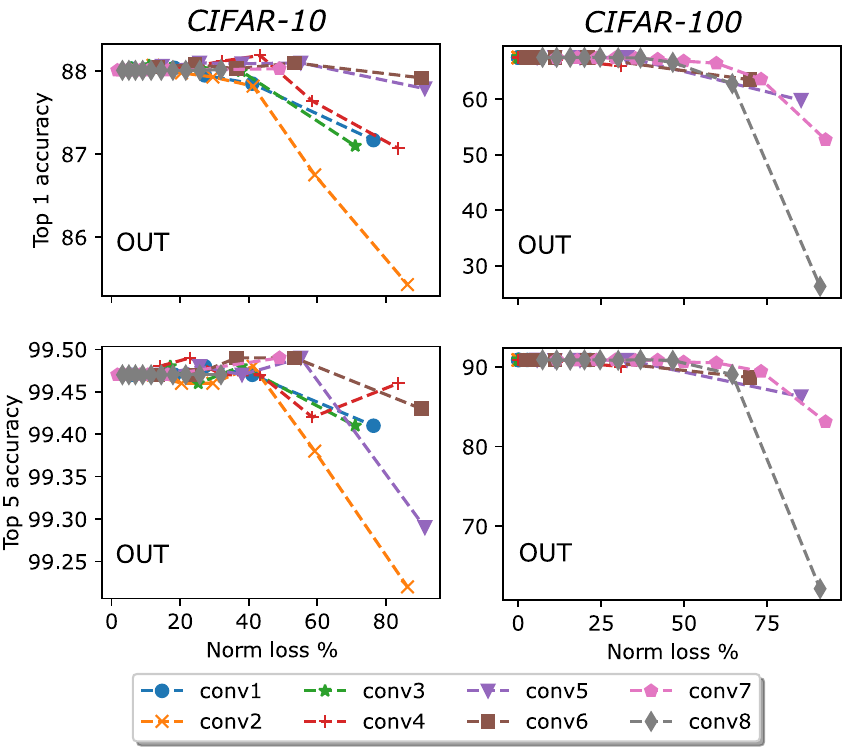}
    \caption{SVD-based truncation across bipartition OUT. The model's accuracy remains robust against up to $50\%$ norm loss in the convolution kernels.}
    \label{fig:single_mode_trunc_out_channels_combined}
\end{figure}

We applied single-mode correlation truncation separately on each mode of the eight convolution kernels. In each case, we replaced the convolution kernel in the model with the corresponding truncated version to obtain a truncated model. We then assessed the truncation's impact on the model's validation accuracy. Even though we report only the impact on validation accuracy, we found similar trends when tracking the impact on training accuracy.

The results of correlation truncation across bipartition OUT are shown in Fig.~\ref{fig:single_mode_trunc_out_channels_combined}. We plot the impact on the model's Top 1 and Top 5 accuracy. We see that truncation up to $50\%$ of the norm has little impact on the model's accuracy. We also see that the model is generally more robust against truncating deeper convolution layers; namely, for a given norm loss, the drop in accuracy is larger for shallower layers.

\begin{figure}
    \centering
    \includegraphics[width=\columnwidth]{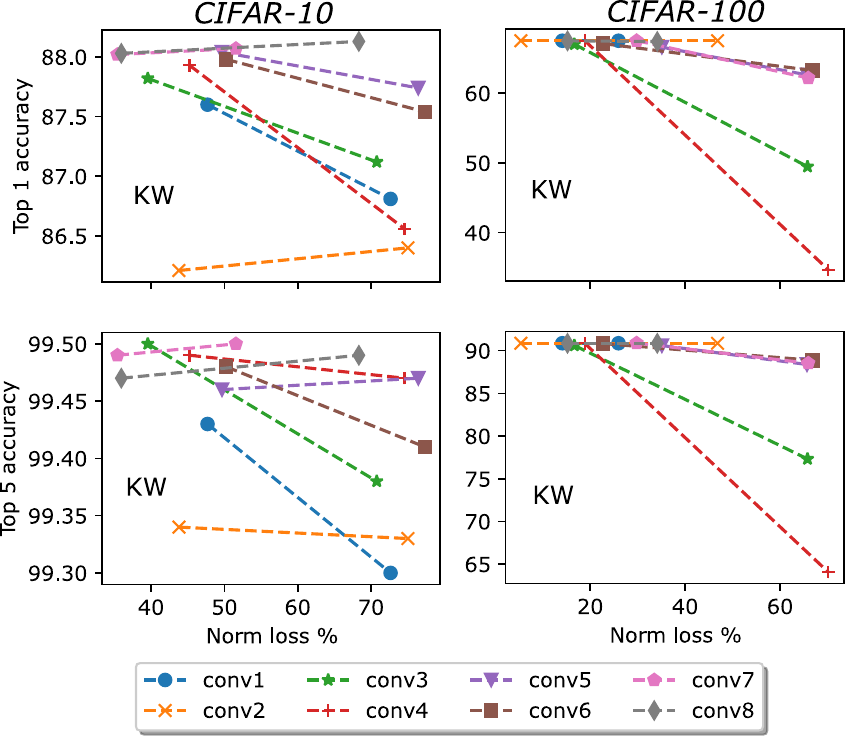}
    \caption{SVD-based truncation across bipartition KW. The truncation impact here is larger than the case for OUT [Fig.~\ref{fig:single_mode_trunc_out_channels_combined}]. This is because bipartition KW has a smaller number of singular values; therefore, each singular value carries a substantial proportion of the total norm.}
    \label{fig:single_mode_trunc_kernel_width_combined}
\end{figure}

Fig.~\ref{fig:single_mode_trunc_kernel_width_combined} shows the impact of single-mode correlation truncation across the bipartition KW. Compared to the truncation across bipartition OUT, discussed above, truncating correlations across the much smaller KW mode has a larger impact on accuracy. This is expected because a smaller number of singular values implies that each singular value captures a larger proportion of the total kernel norm.

\begin{figure}
    \centering
    \includegraphics[width=\columnwidth]{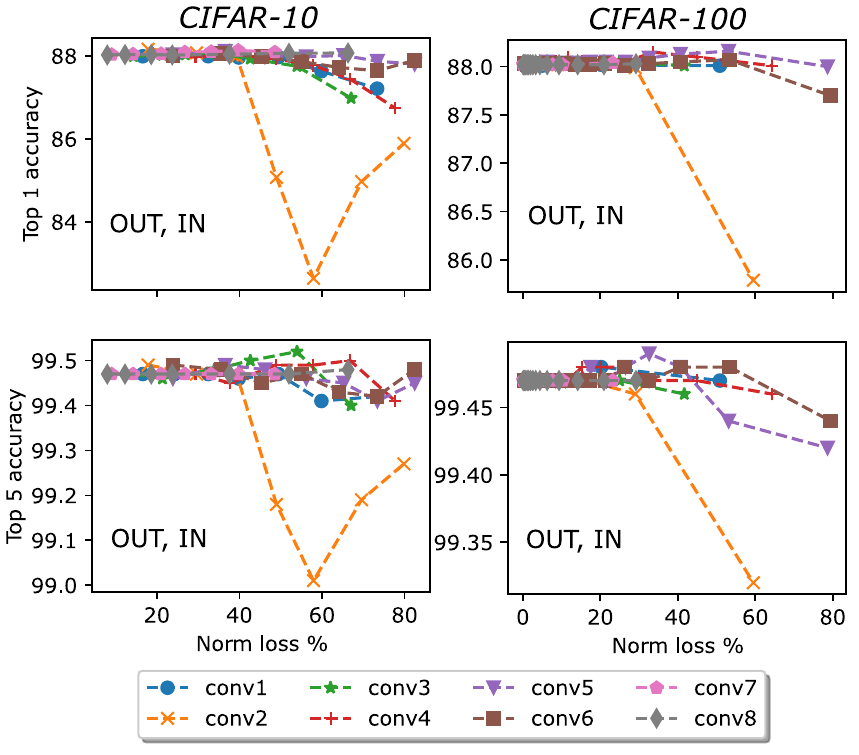}
    \caption{SVD-based truncation across bipartition OUT, IN. The model's accuracy remains robust against up to $50\%$ norm loss in the convolution kernels.}
    \label{fig:pair_mode_trunc_in_out_combined}
\end{figure}

Fig.~\ref{fig:pair_mode_trunc_in_out_combined} shows the results of correlation truncation across bipartition OUT, IN. We find again that the model's accuracy remains resilient to up to $50\%$ norm loss and that deeper layers are more resilient than shallower ones. 

These results indicate that it is possible to compress the convolution kernels across certain cuts without significant loss of accuracy, even though the norm loss can be high.

\begin{figure}
    \centering
    \includegraphics[width=\columnwidth]{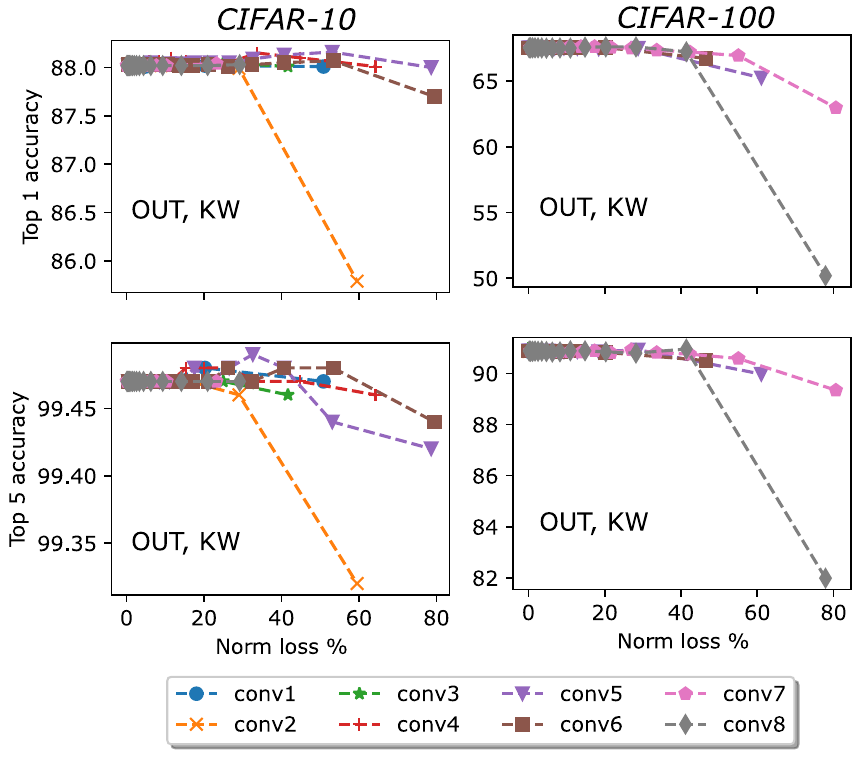}
    \caption{SVD-based truncation across bipartition OUT, KW. The model's accuracy remains robust against up to $40\%$ norm loss in the convolution kernels.}
    \label{fig:single-mode-trunc-kernel_width}
\end{figure}

\begin{figure}
    \centering
    \includegraphics[width=\columnwidth]{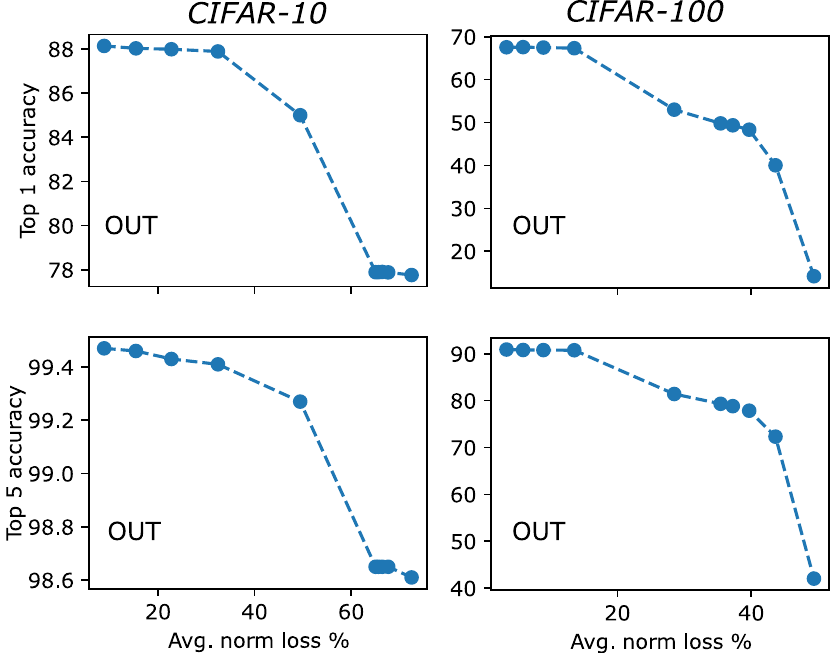}
    \caption{SVD-based truncation across bipartition OUT of all eight convolution layers in parallel. The model's accuracy remains robust against to up to $40\%$ norm loss for CIFAR-10 but only up to $17\%$ norm loss for CIFAR-100.}
    \label{fig:simultaneous_trunc_out_channels_combined}
\end{figure}
\begin{figure}
    \centering
    \includegraphics[width=\columnwidth]{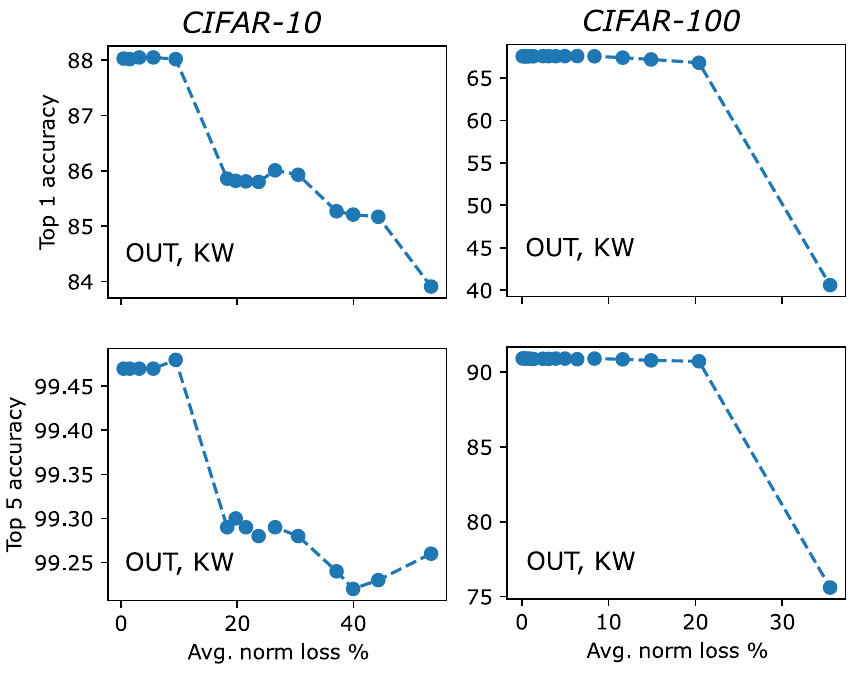}
    \caption{SVD-based truncation across bipartition OUT, KW of all eight convolution layers in parallel. The model's accuracy remains robust against up to $20\%$ norm loss for CIFAR-100 but only up to $10\%$ norm loss for CIFAR-10, the reverse of the trend observed for these datasets in Fig.~\ref{fig:simultaneous_trunc_out_channels_combined}}.
    \label{fig:simultaneous_trunc_pair_out_channels_kernel_width}
\end{figure}
\subsection{Simultaneous correlation truncation across several layers}
We have seen it is possible that a substantial truncation -- up to $50\%$ norm loss in the case of ResNet-50 -- of a single convolution layer may not significantly impact the model's accuracy. Does this robustness result from the large size of ResNet-50 (which help diffuse the impact of single-layer truncations) and/or the presence of residual connections that allow meaningful information to flow past corrupted layers? This doesn't seem to be the case because we observed a comparable robustness against single-layer truncations also in a small four-layer CNN without any residual connections, see Appendix \ref{sec:vanilla}. 

Nonetheless, to rule out the emergence of resilience from a potential bypass of corrupted layers, we also assessed how the model's accuracy diminishes when truncating several layers \textit{concurrently}. We found that robustness does indeed reduce but does not disappear entirely. Figs.~\ref{fig:simultaneous_trunc_out_channels_combined} and \ref{fig:simultaneous_trunc_pair_out_channels_kernel_width} show the results of concurrently truncating the kernel across bipartitions OUT and OUT, KW, respectively, in all eight convolution layers. We observed the following trends.

\textit{Bipartition OUT}: for CIFAR-10, we found that the model's top 1 accuracy remains comparable up to an \textit{average} norm loss of $40 \%$ across all eight convolution layers, while for CIFAR-100, the model preserves top 1 accuracy up to $17 \%$ norm loss.  

\textit{Bipartition OUT, KW}: for CIFAR-10, we found that the model's top 1 accuracy remains comparable up to an average of $10 \%$ norm loss across all eight convolution layers, while for CIFAR-100, the model preserves top 1 accuracy up to $20 \%$ average norm loss.

\subsection{Post-truncation accuracy recovery}
\begin{figure}
    \centering
    \includegraphics[width=\columnwidth]{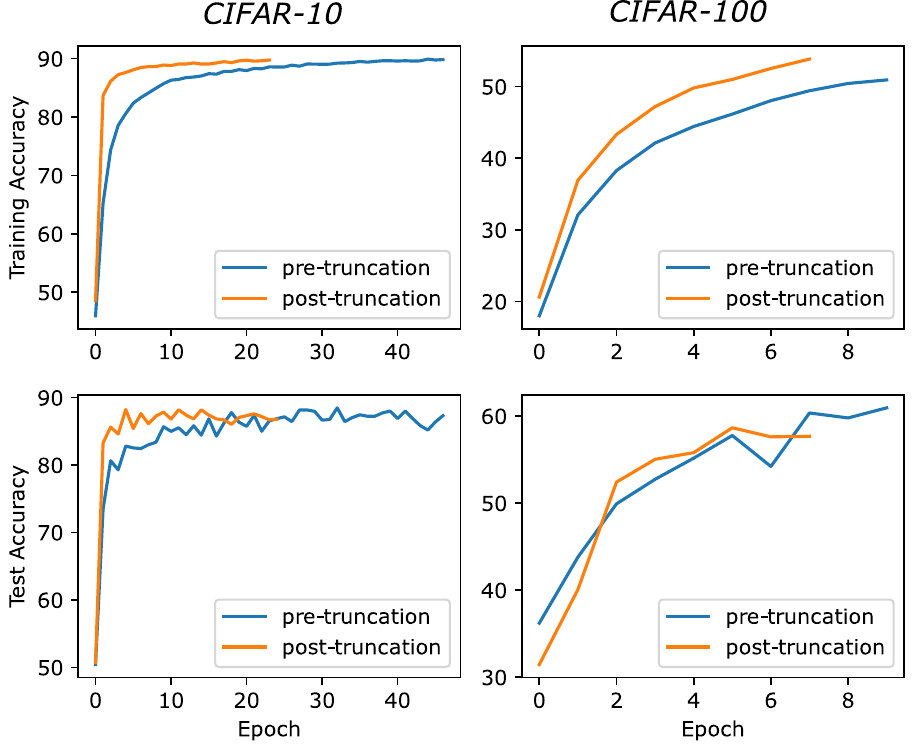}
    \caption{Following an aggressive truncation --- one that resets the accuracy to that of the original untrained model --- the model recovers the pre-truncation accuracy only after a few epochs of re-training.}
    \label{fig:bounce_combined}
\end{figure}

Next, we aggressively truncated all eight convolution layers in parallel across bipartition OUT (Fig.~\ref{fig:simultaneous_trunc_out_channels_combined}) and then re-trained the truncated models to assess how quickly the truncated model regained the accuracy of the original model. The truncation's magnitude was chosen so that the accuracy of the truncated model was comparable to the original \textit{untrained} model. The results for a particular truncation are shown in Fig.~\ref{fig:bounce_combined}. Remarkably, the model recovered the pre-truncation accuracy only after a few training epochs. We tried several other aggressive truncation scenarios and found a similar recovery trend. 

There may be several reasons to account for the quick post-truncation recovery. First, we did not truncate any fully connected layers in these experiments, which might have helped preserve important knowledge in the next. However, it seems unlikely that the presence of the original classifier (fully connected layers) of the trained CNN could help accelerate the re-training of a severely corrupted feature extractor (convolution layers). Note that accuracy dropped significantly post-truncation despite the presence of the original fully connected layers.
Second, the residual connections might have played a role in the quick recovery. However, we observed a quick recovery also in the vanilla CNN model without skip connections; see the bottom plot in Fig.~\ref{fig:cp_and_bounce}. It seems possible, and likely, to us that a fast post-truncation recovery indicates that these particular truncations do not transport the model to a worse minimum. In a recent work Ref.~\cite{MultiverseLLMs}, we also observed a quick post-truncation recovery in a completely different architecture, namely, transformers. The precise mechanism of this recovery should be explored in future work.

\subsection{CP decomposition-based truncation}\label{sec:cptrunc}
\begin{figure}
    \centering
    \includegraphics[width=\columnwidth]{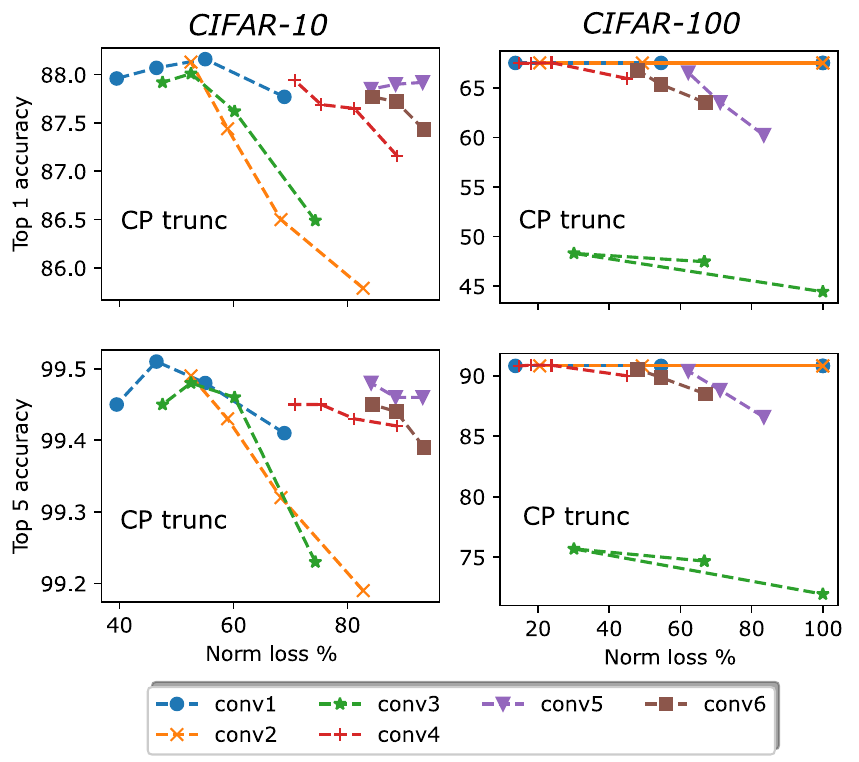}
    \caption{The impact of CP decomposition-based truncation on the model's accuracy. We find that for CIFAR-10, the model remains robust against truncations up to $50\%$ norm loss. This is also the case for CIFAR-100 with the exception of truncations applied on \textit{conv3}, which have a significant impact on the accuracy, possibly due to the fact that  \textit{conv1} and  \textit{conv2} have vanishing norm in this case, increasing \textit{conv3}'s contribution to the total accuracy.}
    \label{fig:cp-trunc}
\end{figure}

We also carried out layer-wise CP decomposition-based truncation of convolution kernels. The results are shown in Fig.~\ref{fig:cp-trunc}. For both CIFAR-10 and CIFAR-100, we found that the model is robust against truncating up to $50\%$ (in norm) of the convolution kernel, with the exception of truncating \textit{conv3} for CIFAR-100, which appears to diminish the accuracy strongly. One possible explanation for this outlier could be that the previous two convolution layers, \textit{conv1} and \textit{conv2}, have vanishing norm, so truncating \textit{conv3}, which essentially is the first layer to receive the input data features, impacts the accuracy more strongly.

\section{Conclusions} 
\label{sec:conclude}
We have demonstrated that dense (untensorized) CNNs can show remarkable robustness against truncating correlations between the weights stored inside convolution kernels. 

Training a tensorized model is, in principle, different from training the corresponding dense model in that the loss surface (and the gradient computation during backpropagation) can be very different in the two cases. The fact that dense CNNs are robust against correlation truncation suggests that the practical advantages of training a tensorized version of the model do not entirely stem from these differences but perhaps because the intrinsic organization of information in CNNs makes them more amenable to tensorization. In other words, our results suggest that tensorized formats are more \textit{optimal} representations of convolution kernels, since the correlations between trained kernel weights can be compressed without sacrificing accuracy significantly.

Our impact assessment strategy also has practical applications. First, it can be applied to compress a trained CNN optimally -- the kernels that have a greater impact on accuracy should be compressed more mildly. Second, assessing the impact of truncations on \textit{dense} models can provide useful information for fixing hyperparameters involved in designing and training \textit{tensorized} CNN models. Tensorized models can contain a huge number of hyperparameters, such as the tensor network geometry of the models, the number of tensors in each layer, and the dimensions of the various tensor indices. We can try to fix some of these hyperparameters by, for instance, \textit{partially} training a dense prototype CNN and then truncating it -- keeping the impact on accuracy below a threshold -- to obtain a tensorized CNN model that can then be fully trained for the task at hand. 

Another interesting effect that deserves further study is the quick rebound in accuracy when re-training after correlation truncation. Previously, we also reported a quick post-truncation recovery when truncating and re-training transformers in the context of large language models \cite{MultiverseLLMs}. For instance, can this effect be understood as a consequence of a particular feature of the loss surface?

The robustness against correlation truncation reported here for convolution layers and CNNs should also apply to other layers, e.g., fully connected layers and attention layers, and other neural architectures such as multi-layer perceptrons, transformers, etc. We believe that understanding the origin of robustness against correlation truncation and the post-truncation recovery are crucial to understanding the efficacy of tensorized neural networks for deep learning.

\section*{Acknowledgements}
S.S.J. acknowledges the support from the Institute for Advanced Studies in Basic Sciences (IASBS), Donostia International Physics Center (DIPC), and Multiverse Computing.

\paragraph{Author contributions}
All authors contributed equally to the manuscript.


\begin{appendix}
\numberwithin{equation}{section}

\section{Brief review of tensor networks}\label{app:A}

A \textit{tensor} is a multi-dimensional array of numbers. (For our purposes, we restrict attention to real-valued tensors.) The numbers inside a tensor are accessed by specifying values of a set of indices (array coordinates). Familiar examples of tensors include scalars, vectors, one-index tensors, and matrices, see Fig.~\ref{fig:tensors} (i), (ii).

Matrices are two-index tensors where the two indices enumerate the rows and columns. For instance, for a $2 \times 3$ matrix $M_{ij}$, the two indices take values $i=\{0,1\}$ and $j=\{0,1,2\}$. (We follow the Python programming language's index numbering convention which begins at 0.)

The \textit{size} or \textit{dimension} of an index is the number of values it can assume. We denote the dimension of an index by $|\,\,|$, e.g., for the $2 \times 3$ matrix $M$ we denote $|i|=2$  and $|j|=3$.

The \textit{size} of a tensor is the number of components it has, equal to the product of the dimensions of each of its indices. For instance, $|M| = |i||j|$.

In this paper, we follow a graphical calculus for tensors that is used widely in quantum many-body physics \footnote{A graphical calculus for tensors was first popularized by Penrose. The current graphical calculus of quantum tensor networks can be formalized as string diagrams in a suitable category.}. The graphical representation of tensors and some basic tensor operations are illustrated in Fig.~\ref{fig:tensors}. Panels (v)-(vii) in the figure introduce special tensors we use later to describe CNNs.

A tensor network (TN) is a collection of tensors that can be multiplied or contracted together according to a specified network; see Fig.~\ref{fig:tensors}(x). We distinguish two types of indices in a tensor network. A \textit{bond} index connects two tensors in the network, while an \textit{open} index is connected only to a single tensor. 

Given a TN, we can obtain a single tensor by contracting or multiplying the tensors according to the network. Tensor contraction is a generalization of the multiplication of matrices, Fig.~\ref{fig:tensors}(viii), to higher-dimensional tensors. It is carried out by summing over the bond indices and multiplying the respective components of the tensors in the network. A popular implementation of tensor contraction is the \textbf{einsum} function within the NumPy Python library. For example, the tensor contraction depicted in 
Fig.~\ref{fig:tensors}(x), and detailed in Fig.~\ref{fig:cost}(i), can be implemented in Python as:
\begin{lstlisting}[language=Python]
import numpy as np
t = np.einsum('iga, ab, gkjb -> ikj')
\end{lstlisting}

\begin{figure}
    \centering
    \includegraphics[width=0.7\columnwidth]{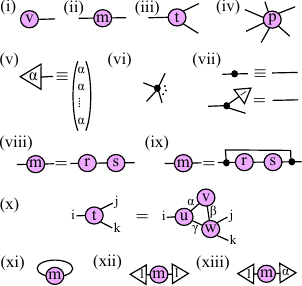}
    \caption{Graphical representations of elementary tensors and tensor operations. (i) A vector $v_i$. (ii) A matrix $m_{ij}$. (iii) A 3-index tensor $t_{ijk}$. (iv) A 6-index tensor $p_{ijklmn}$. (v) Specialized representation of a vector whose components are all $\alpha$. (vi) A copy (or delta) tensor, whose components are equal to one for equal value of all indices, and 0 otherwise. (vii) (Top) The copy tensor is a higher-dimensional generalization of the identity matrix; we depict the identity matrix (which is a 2-index copy tensor) as a straight line. (Bottom) A general property of the copy is that contraction with a vector of all ones equates to a copy tensor with one fewer index. (viii) Matrix multiplication $m = r \times s$. (ix) Elementwise product (Hadamard product) of matrices $r$ and $s$. (x) A more general tensor contraction of 3 tensors $u,v,w$ that results in tensor $t$. (xi) Trace of matrix $m$. (xii) The sum of elements of matrix $m$. (xiii) $\alpha$ times the sum of elements of matrix $m$.}
    \label{fig:tensors}
\end{figure}

\subsection{Computational cost of a TN contraction}\label{sec:TNcost}
When implementing a tensor network contraction, a central task is estimating its computational cost, namely, the total computational time required for carrying out the elementary number multiplications and additions.

We can estimate the total number of elementary operations involved in TN contraction by means of a simple rule:

\textbf{Rule 1:} The total number of elementary operations involved in TN contraction is proportional to the product of the dimensions of all the open and bond indices involved in the contraction. 

For example, applying this rule to the contraction shown in Fig.~\ref{fig:tensors}(x), we can estimate that it incurs a computational cost proportional to $|i||j||k||\alpha||\beta||\gamma|$.

\section{Cost analysis for Tucker decomposed convolutions}\label{sec:cost}
In this section, we carry out a cost analysis for Tucker-decomposed convolutions as an example. (Costs for CP-decomposed convolutions can be estimated analogously.) By decomposing the convolution kernel via Tucker decomposition, we can save significantly on the memory required to store the kernel and in the computational time required to perform the convolution during training and inference. We describe how a tensor network description of tensorization provides a convenient way to estimate these gains.

\subsection{Memory compression} 
First, let's consider the memory compression. The size of a dense convolution kernel $K$ (Eq.~\ref{eq:kerneldef}) is proportional to the total number of components, which equals 
\begin{equation}
\begin{split}
&~|x||y||||| \\
&= XYC^{\mbox{\tiny in}}C^{\mbox{\tiny out}}. 
\end{split}
\end{equation}
On the other hand, the total number of components in the Tucker decomposition of $K$ equals 
\begin{equation}
\begin{split}
&~|x||\alpha| + |y||\beta| + |||\gamma| + |||\delta| \\
&= X|\alpha| + Y|\beta| + C^{\mbox{\tiny in}}|\gamma| + C^{\mbox{\tiny out}}|\delta|. 
\end{split}
\end{equation}
In practical applications, we find that the Tucker ranks can be significantly smaller than the dimensions of the original kernel, which often results in a substantial compress ratio CR given by:
\begin{align}
\mbox{CR} = \frac{XYC^{\mbox{\tiny in}}C^{\mbox{\tiny out}}}{X|\alpha| + Y|\beta| + C^{\mbox{\tiny in}}|\gamma| + C^{\mbox{\tiny out}}|\delta|}.
\end{align}
\textbf{Example 1.} Let's assume a kernel size $X=3, Y=3$, number of input channels $|| = C^{\mbox{\tiny in}} = 256$, and number of output channels $|| = C^{\mbox{\tiny out}} = 384$. (These numbers correspond to the dimensions of a convolution layer inside e.g. AlexNet.). We fix the Tucker ranks $|\alpha|=|\beta|=3$, and for simplicity, let us assume that the remaining two ranks are also equal (but variable), $|\gamma|=|\delta| = \chi$. The compression ratio CR for several values of $\chi$ is listed in the table below:
\begin{table}[h!]
\centering
 \begin{tabular}{||c|c||} 
 \hline
 Tucker Rank ($\chi$) & Compression Ratio (CR) \\ [0.5ex] 
 \hline\hline
 200 & 7$\times$ \\ 
 150 & 9$\times$ \\
 100 & 14$\times$ \\
 50 & 28$\times$ \\ 
 20 & 69$\times$ \\ [1ex] 
 \hline
 \end{tabular}
\end{table}

\subsection{Convolution speedup} 
\begin{figure}
    \centering
    \includegraphics[width=0.7\columnwidth]{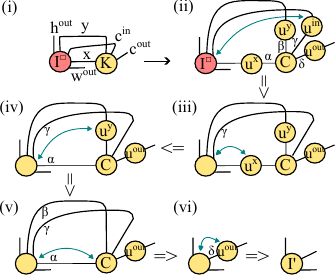}
    \caption{Tucker-decomposed convolutions can be applied as a sequence of 6 pairwise tensor contractions, (i)-(vi). In each step, the double-arrowed curve points to the pair of tensors that are contracted.}
    \label{fig:cost}
\end{figure}

Next, let us compare the computational time required to apply a dense vs. Tucker-decomposed convolution. We can estimate the cost easily when viewing convolutions as tensor network contractions. See Fig.~\ref{fig:cost}. The cost for a dense contraction, panel (i) in Fig.~\ref{fig:cost}, is proportional to the dimensions of all the indices involved in the contraction,
\begin{align}\label{eq:denseconvcost}
\begin{split}
\mbox{Dense Cost} &= |h^{\mbox{\tiny out}}||w^{\mbox{\tiny out}}||x||y||||| \\
&= H^{\mbox{\tiny out}} W^{\mbox{\tiny out}} X Y C^{\mbox{\tiny in}} C^{\mbox{\tiny out}}.
\end{split}
\end{align}
Let's compare this with the cost of applying the kernel given in the Tucker-decomposed format. In this case, the entire operation can be applied as a sequence of pairwise contractions as depicted in Fig.~\ref{fig:cost}. First, the mode matrices $U^X, U^Y,$ and $U^{\mbox{\tiny in}}$ are multiplied with the patch image tensor $I^{\square}$ in an optimal order to compress the corresponding modes of the latter. The optimal contraction order corresponds to first contracting the mode matrix on the largest input dimension, followed by contracting mode matrices in order of decreasing dimensions. Here we assume that $C^{\mbox{\tiny in}} > X \geq Y$. The total cost of thus absorbing the mode matrices $U^{\mbox{\tiny in}}, U^X, $ and $U^Y$, in that order, into the image is
\begin{align}\label{eq:cost1}
\begin{split}
\mbox{Cost 1}=&H^{\mbox{\tiny out}} W^{\mbox{\tiny out}} (XY C^{\mbox{\tiny in}}|\gamma|~+ \\ 
& XY|\gamma||\alpha|~+~|\alpha|~Y~|\gamma||\beta|) .
\end{split}
\end{align}
Next, we contract the compressed input image with the core tensor $C$. The cost of this contraction is
\begin{align}\label{eq:cost2}
\mbox{Cost 2} = H^{\mbox{\tiny out}} W^{\mbox{\tiny out}} |\alpha| |\beta| |\gamma| |\delta|
\end{align}
Finally, we multiply in the mode matrix $U^{\mbox{\tiny out}}$, incurring a cost of 
\begin{align}\label{eq:cost3}
\mbox{Cost 3} = H^{\mbox{\tiny out}} W^{\mbox{\tiny out}}|\delta| C^{\mbox{\tiny out}}.
\end{align}
The total cost for applying the convolution is obtained by adding the costs listed in Eqs.~\ref{eq:cost1}-\ref{eq:cost3}, 
\begin{equation}
\mbox{Tucker Cost = Cost 1 + Cost 2 + Cost 3},
\end{equation}
which can be substantially smaller than the cost for applying a dense convolution, Eq.~\ref{eq:denseconvcost}, resulting in a speedup equal to (Dense Cost)/(Tucker Cost). Note again that, in practice, the Tucker ranks $|\alpha|, |\beta|, |\gamma|, |\delta|$ are relatively smaller than the dimensions of the input patch image tensor. Thus, Tucker-decomposed convolutions are, in practice, faster than dense convolutions, implying that TCNNs have faster inference and per-iteration times during training.

\textbf{Example 2.} Consider a convolution with dimensions and Tucker ranks equal to those considered in Example 1 that acts on the input image (specified as a patch image tensor) to produce an image of size $H^{\mbox{\tiny out}} = 50,  W^{\mbox{\tiny out}} = 50$. The speedups obtained by applying the Tucker-decomposed convolution vs. a dense convolution by varying $\chi$ are summarized in the table below:
\begin{table}[h!]
\centering
 \begin{tabular}{||c|c||} 
 \hline
 Tucker Rank ($\chi$) & Speed-up \\ [0.5ex] 
 \hline\hline
 200 & 1.4$\times$ \\ 
 150 & 2$\times$ \\
 100 & 3$\times$ \\
 50 & 6.7$\times$ \\ 
 20 & 17.3$\times$ \\ [1ex] 
 \hline
 \end{tabular}
\end{table}

\section{Truncation results for a small vanilla CNN}\label{sec:vanilla}
As a simpler demonstration of robustness against correlation truncation, we ran the experiments described in the main text for ResNet-50 on a simple four-layer CNN trained for image classification on the CIFAR-10 dataset. 

\begin{figure*}
\centering
    \centering
    \includegraphics[width=7cm]{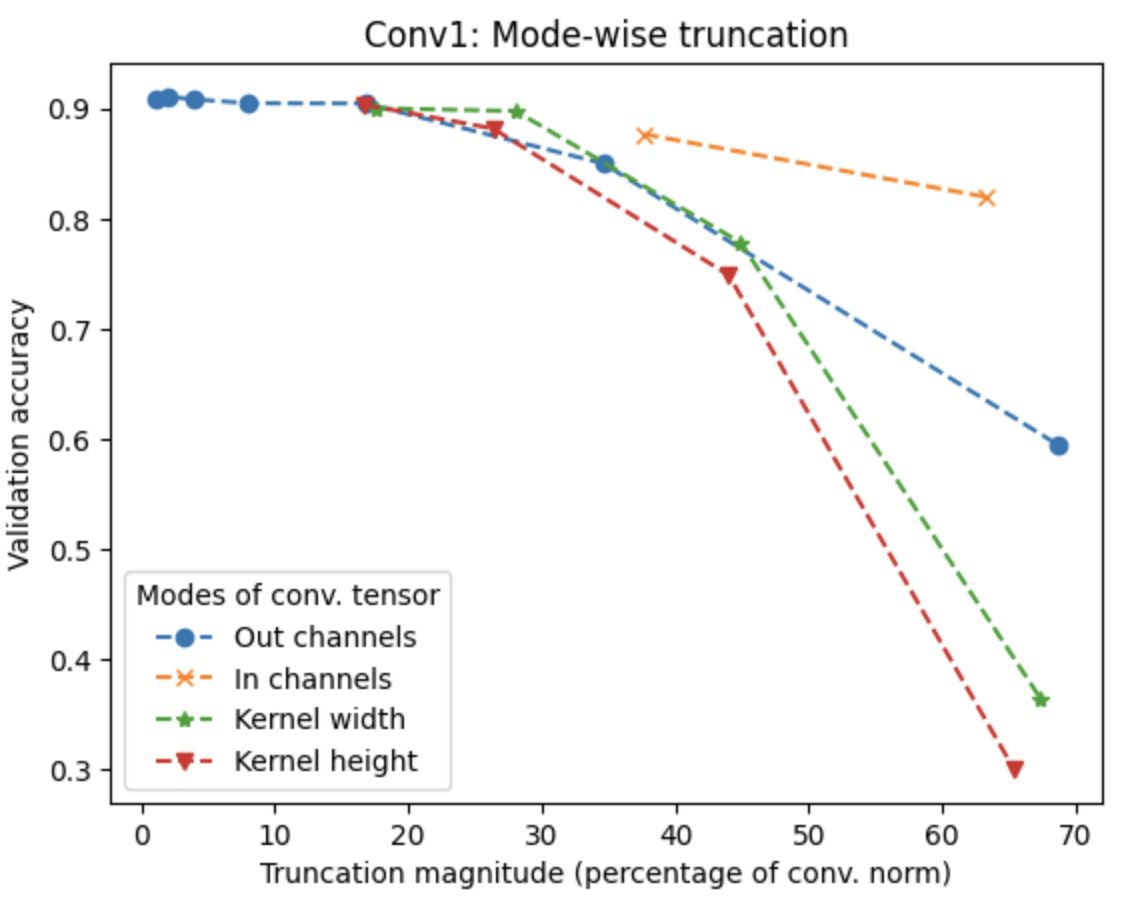}
    \includegraphics[width=7cm]{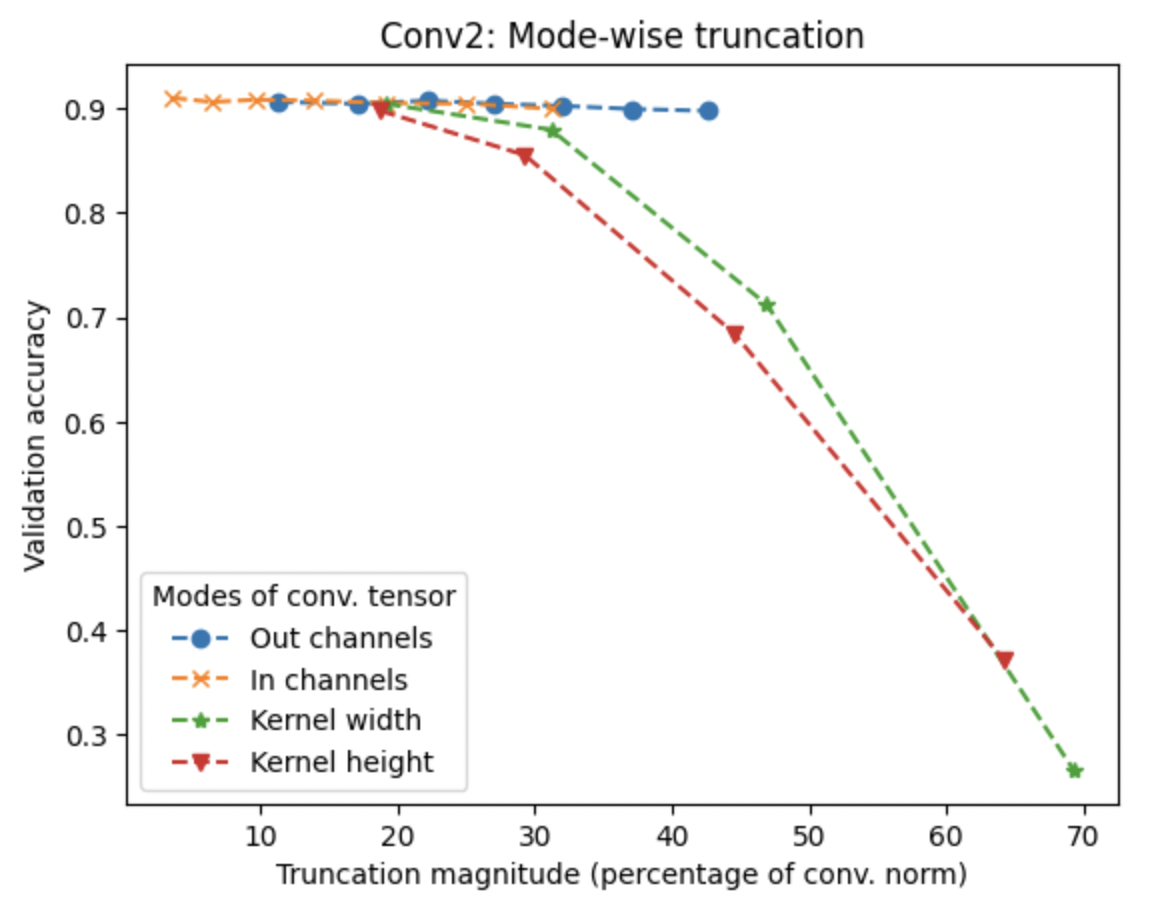}\\
    \includegraphics[width=7cm]{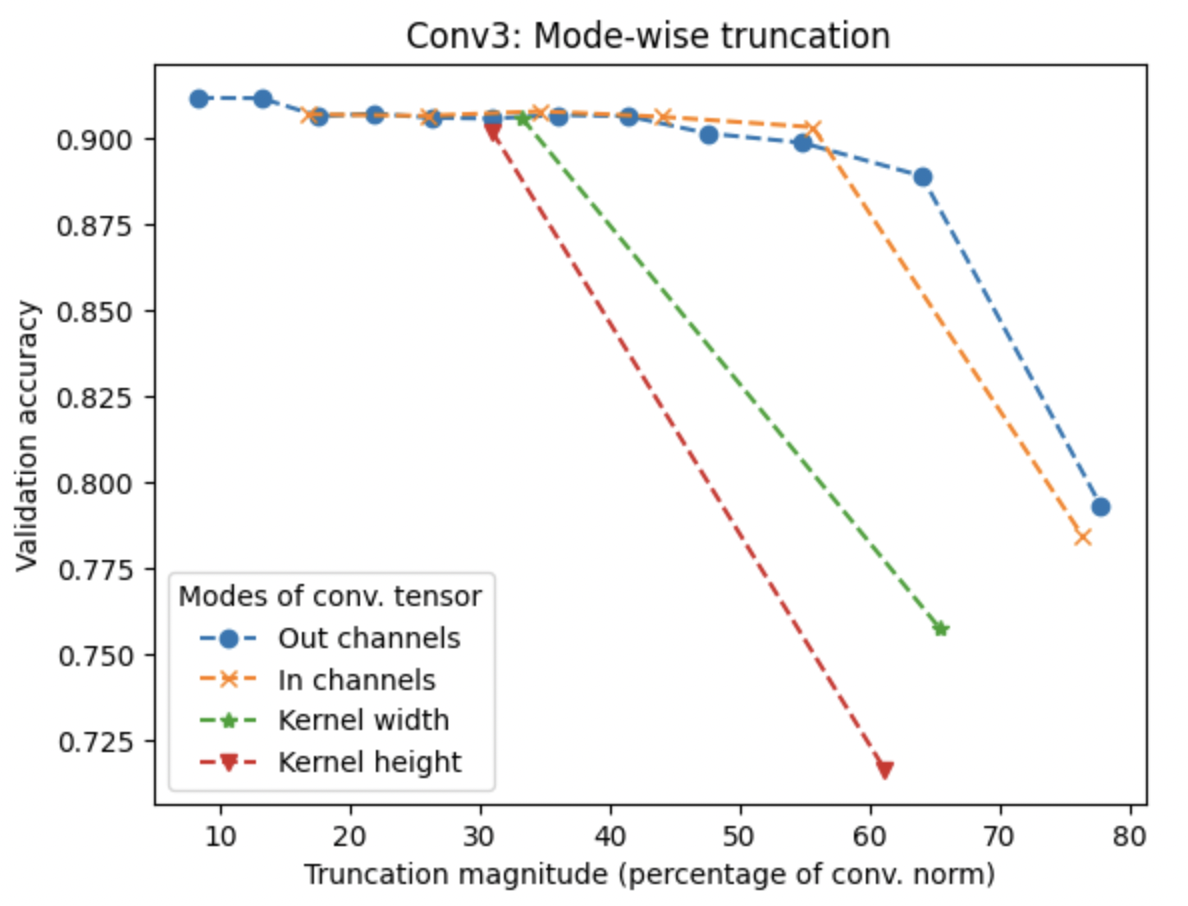}
    \includegraphics[width=7cm]{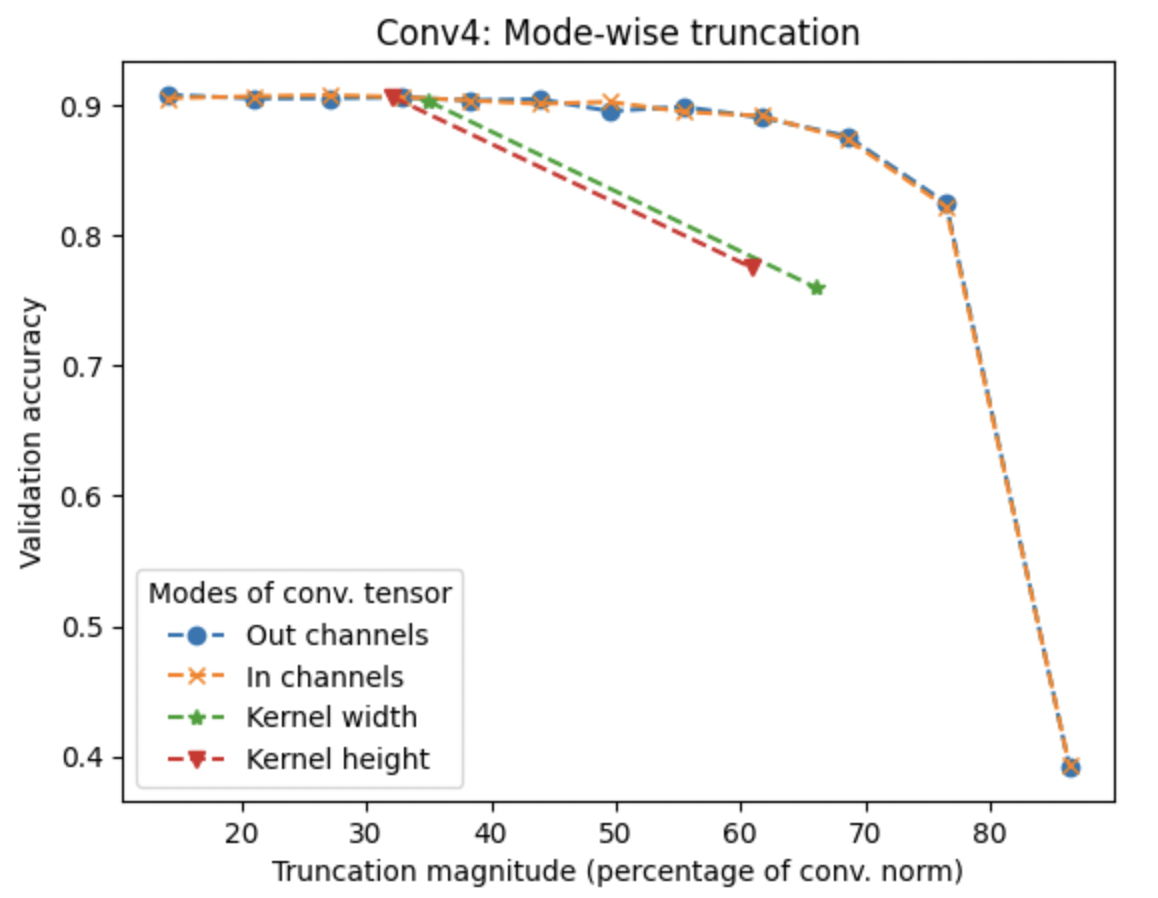}
\caption{Impact of single-mode correlation truncation for each of the four modes of the convolution kernel at various depths in the CNN. Shown here is the impact on validation accuracy. (A similar trend was also found for training accuracy.) Each point in the plot corresponds to discarding $\phi$ number of singular values (see, for instance, discussion around Eq.~\ref{eq:trunc}), which results in loss of norm of the kernel (plotted along the x-axis) and in loss of validation accuracy of the model (plotted along the y-axis). For \textit{conv1}, (the first convolution kernel), we chose the following values of $\phi$ for each of the four modes: $\{10:10:70, 1:1:2, 1:1:4, 1:1:4\}$, where we have used Python-like range notation,  start: step size: end (end is included), to specify a range of values. For instance, $1:1:4 = [1,2,3,4]$ The total number of singular values across a bipartition is upper bounded by the smaller dimensions of the modes across the bipartition. Thus, some lines in the plot have fewer points than others. For \textit{conv2}, $\phi \in \{10:10:70, 10:10:70, 1:1:4, 1:1:4\}$. For \textit{conv3}, $\phi \in \{20:20:240, 20:20:120, 1:1:2, 1:1:2\}$. For \textit{conv4}, $\phi \in \{20:20:240, 20:20:240, 1:1:2, 1:1:2\}$.}\label{fig:trunc_single_modes}
\end{figure*}

The architecture of the model is as follows. The feature extractor consists primarily of 4 convolution layers (each a 2d convolution) with the specifications listed in the table below.
\begin{table}[h]
\centering
 \begin{tabular}{||c|c|c|c|c|} 
 \hline
 CNN Layer & in channels & out channels & kernel size & padding \\ [0.5ex] 
 \hline\hline
 conv1 & 3 & 128 & 5 & 2 \\ 
 conv2 & 128 & 128 & 5 & 2 \\
 conv3 & 128 & 256 & 3 & 1 \\
 conv4 & 256 & 256 & 3 & 1 \\ [1ex] 
 \hline
 \end{tabular}
\end{table}
Each convolution layer consists of a convolution (from the above list), followed by batch normalization and a ReLu activation. A pooling and dropout is applied after conv2 and then after conv4. The classifier component of the CNN consists of 3 fully connected layers, each consisting of a linear map, batch normalization, and a ReLu activation. We also applied dropout after the first and second fully connected layers. The model consists of more than 18 million trainable parameters. 



For training, we used the cross entropy loss function and the Adam optimizer with a learning rate equal to $0.01$ and the beta values (coefficients used for computing running averages of gradient and its square) equal to $(0.9, 0.999)$. We also applied a learning rate scheduler, which lowered the learning rate by a factor of $0.1$ whenever the loss began to plateau.


We trained this model several times with different initializations and obtained an average training loss of $0.1\%$ and validation accuracy of approximately $90\%$. 

\begin{figure*}
\centering
    \includegraphics[width=7cm]{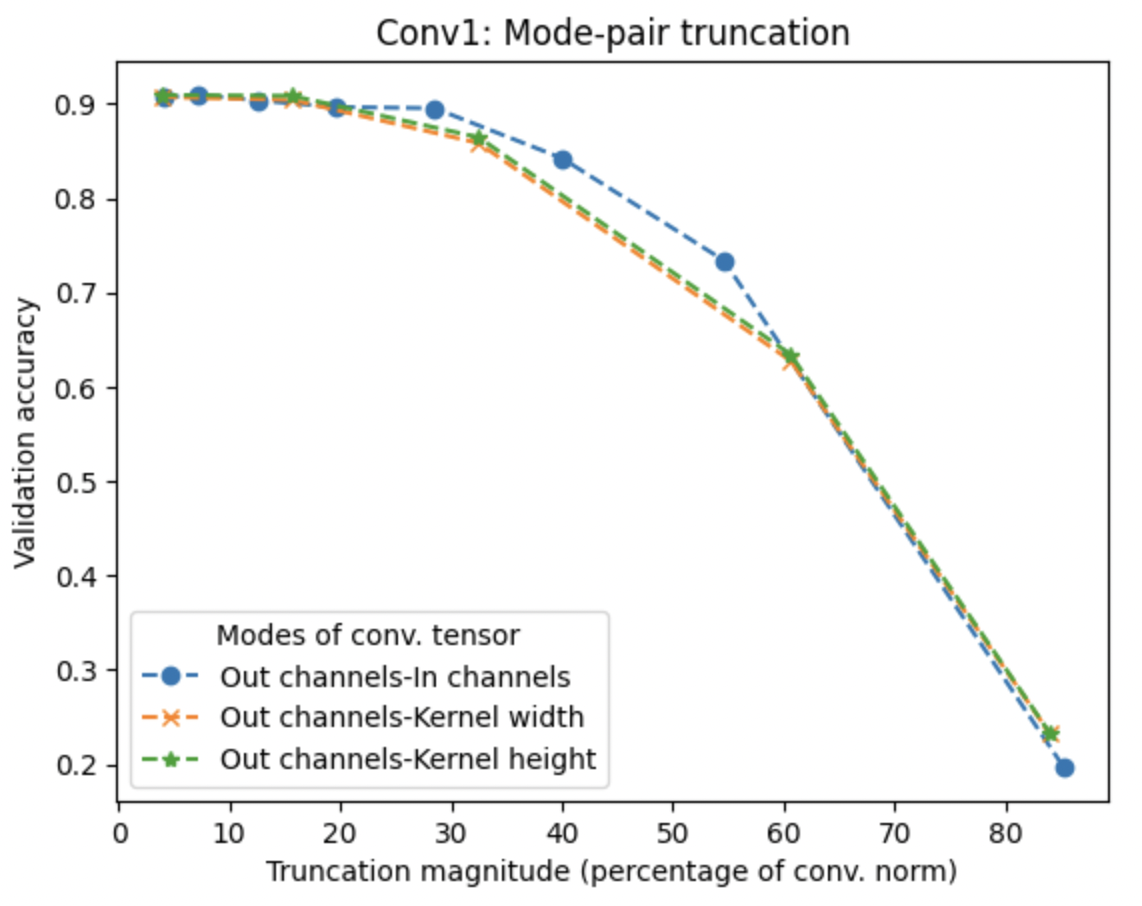}
    \includegraphics[width=7cm]{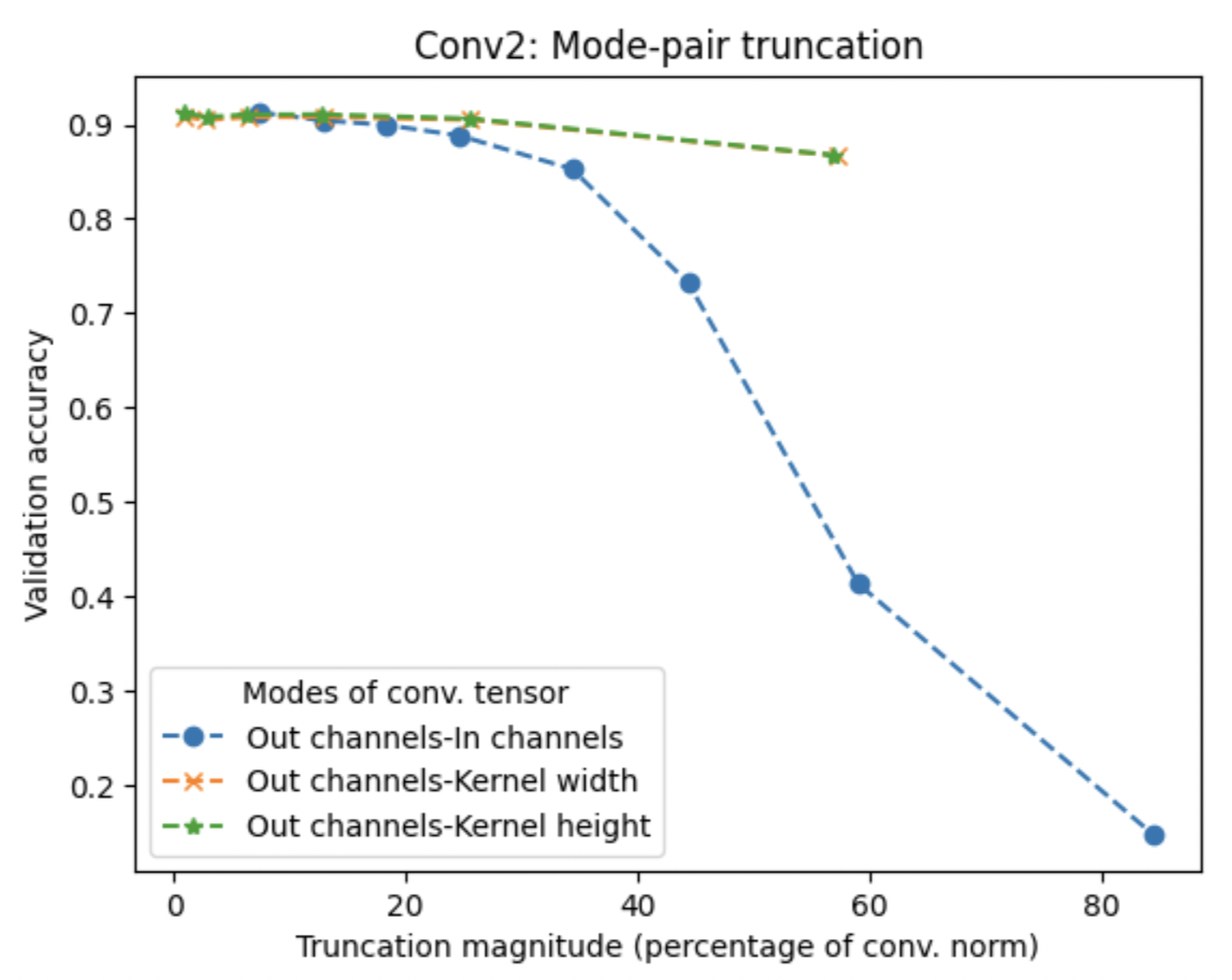}\\
    \includegraphics[width=7cm]{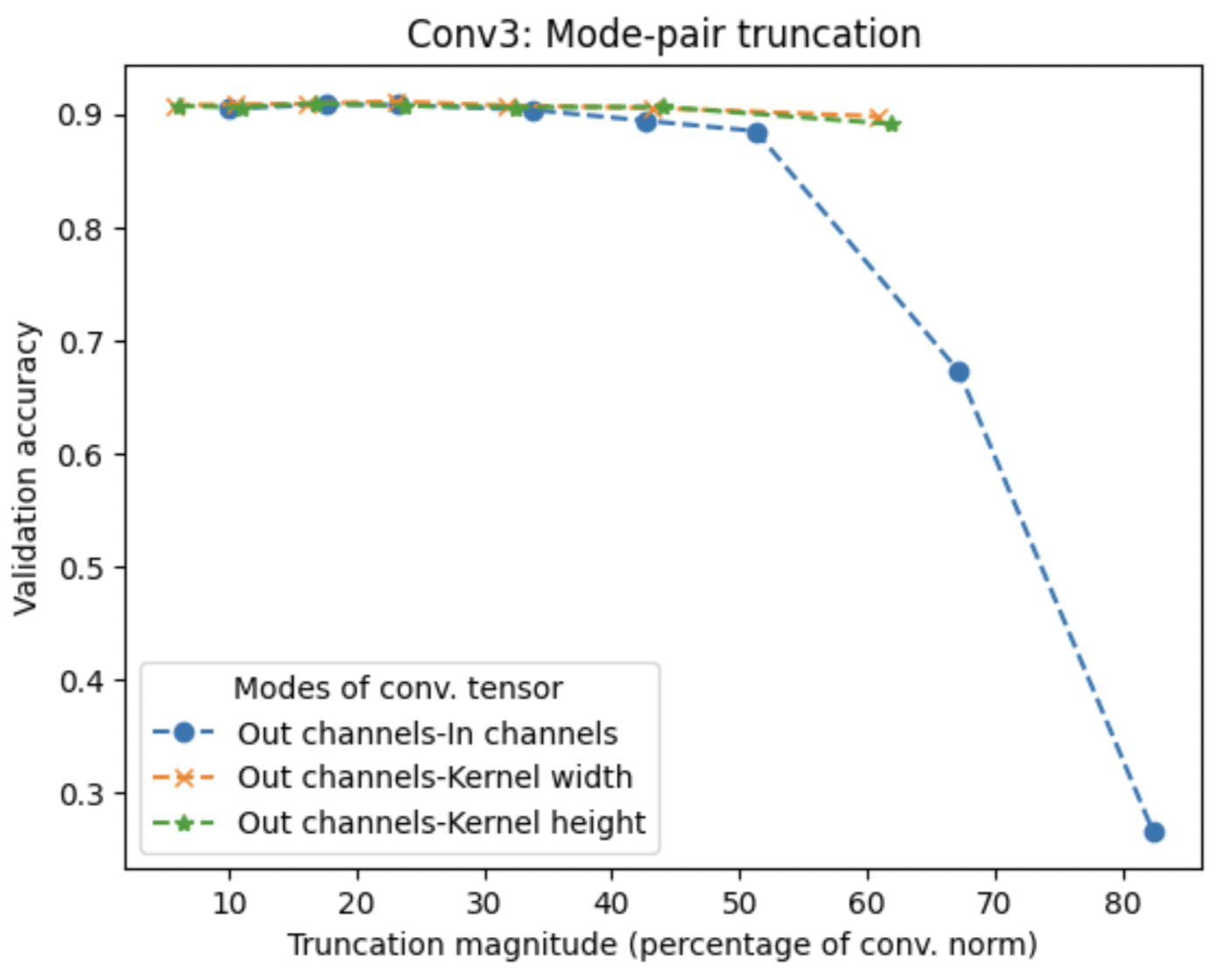}
    \includegraphics[width=7cm]{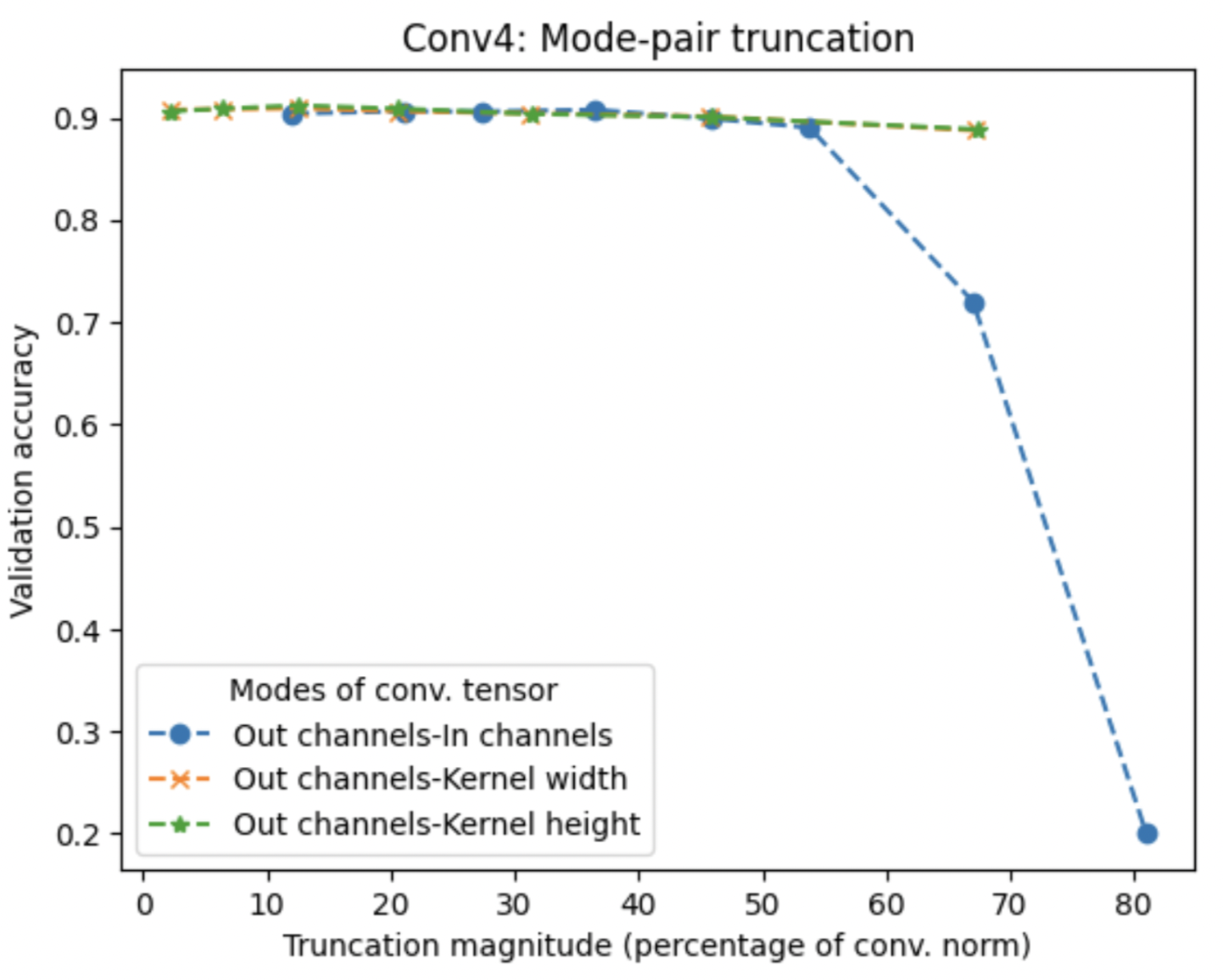}
\caption{Impact of two-mode correlation truncations for convolution kernel at various depths in the CNN. Shown here is the impact on validation accuracy. (A similar trend was also found for training accuracy.) Each point in the plot corresponds to discarding $\phi$ number of singular values (see, for instance, discussion around Eq.~\ref{eq:trunc}), which results in loss of norm of the kernel (plotted along the x-axis) and in loss of validation accuracy of the model (plotted along the y-axis). For \textit{conv1}, $\phi \in \{3:3:24, [3,6,9,12,14], [3,6,9,12,14]\}$. (Range notation is introduced in the caption of Fig.~\ref{fig:trunc_single_modes}.) For \textit{conv2}, $\phi \in \{3:3:24, 100:100:600, 100:100:600\}$. For \textit{conv3}, $\phi \in \{1:1:8, 50:50:350, 50:50:350\}$. For \textit{conv4}, $\phi \in \{1:1:8, 100:100:700, 100:100:700\}$.}\label{fig:mode_pair_trunc}
\end{figure*}

\begin{figure}
\centering
    \includegraphics[width=7cm]{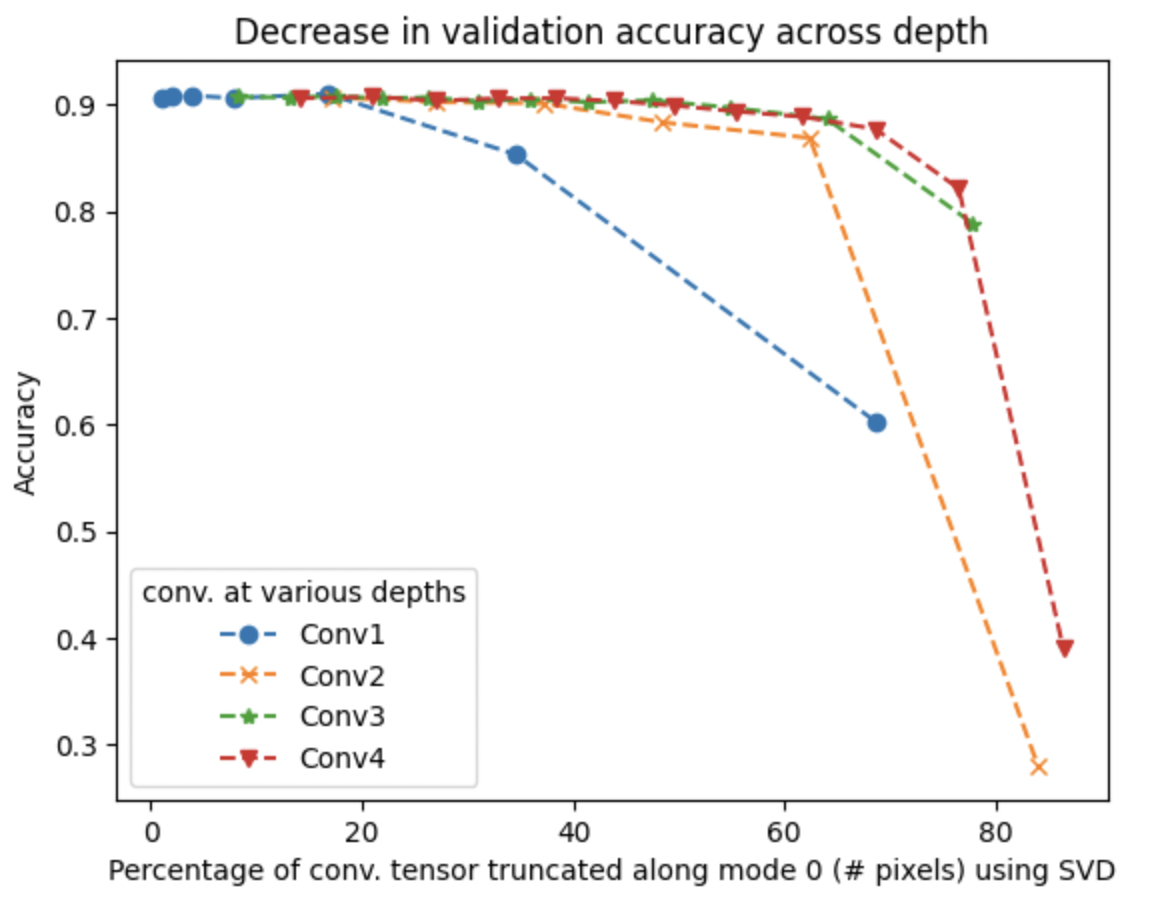}\\
    \includegraphics[width=7cm]{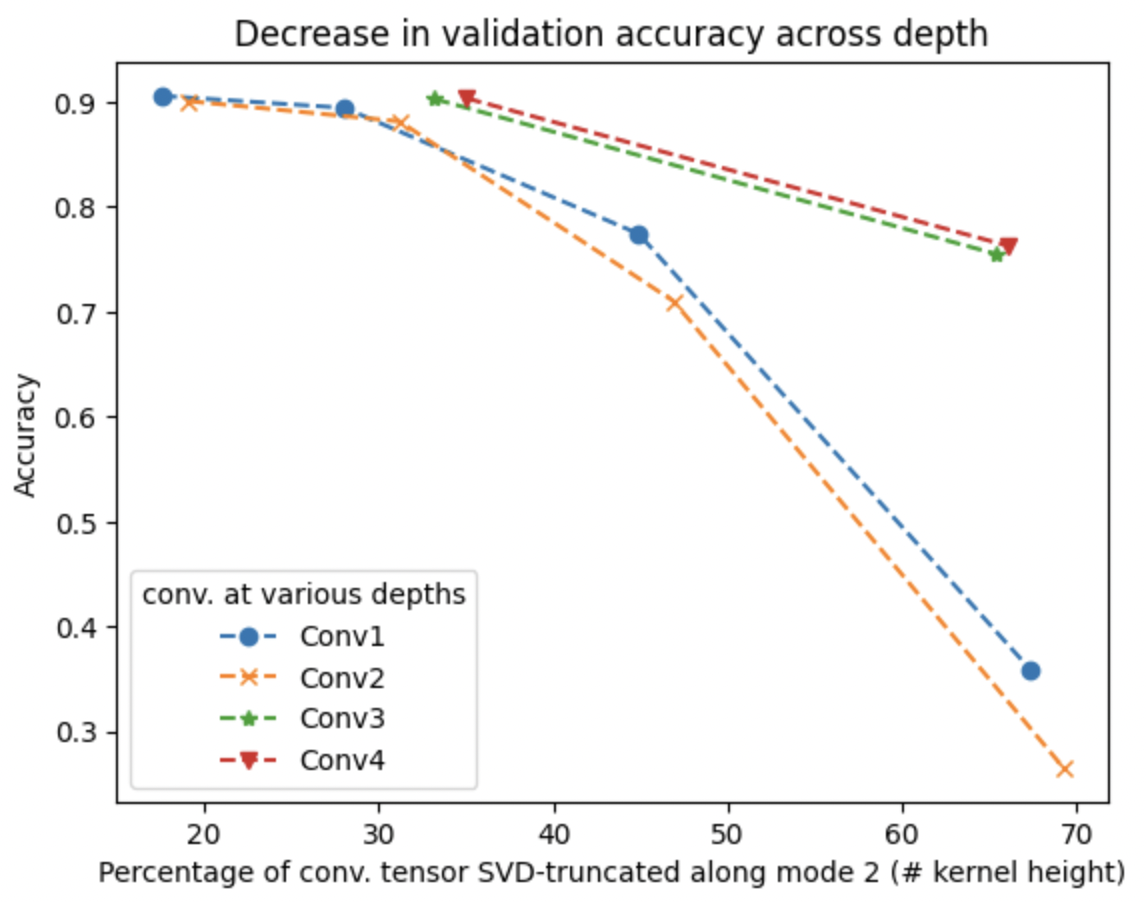}\\
    \includegraphics[width=7cm]{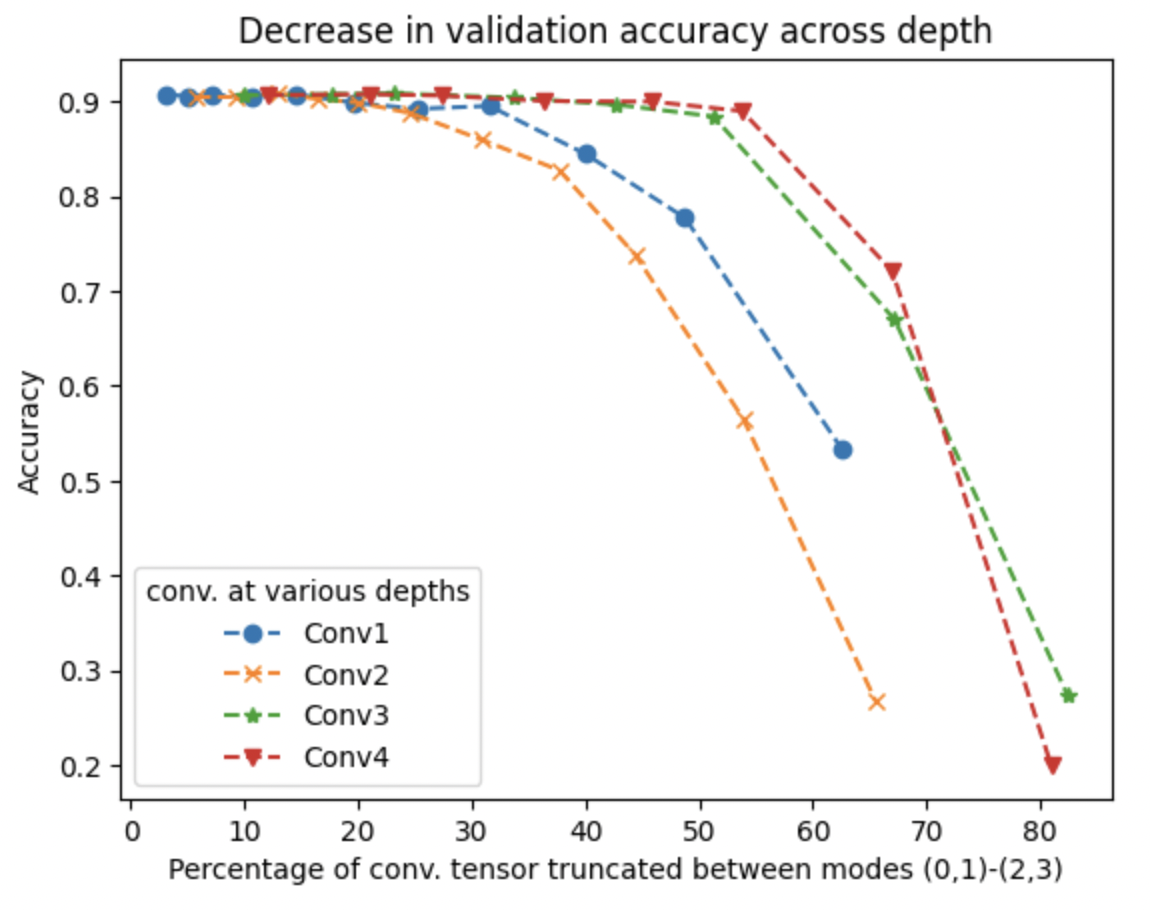}
\caption{Impact of correlation truncation vs depth. (Top Left) Single-mode correlation truncation for out-channel mode. Values of $\phi$ for the 4 layers are $\{10:10:70, 20:20:120, 20:20:240, 20:20:240\}$. (Top Right) Single-mode correlation truncation for kernel width mode. Values of $\phi$ for the 4 layers are $\{1:1:4, 1:1:4, 1:1:2, 1:1:2\}$. (Bottom) Two-mode correlation truncation for combined kernel width and kernel height modes. Values of $\phi$ for the 4 layers are $\{2:2:22, 2:2:22, 1:1:8, 1:1:8\}$.}\label{fig:depth_scaling}
\end{figure}

\begin{figure}
\includegraphics[width=7cm]{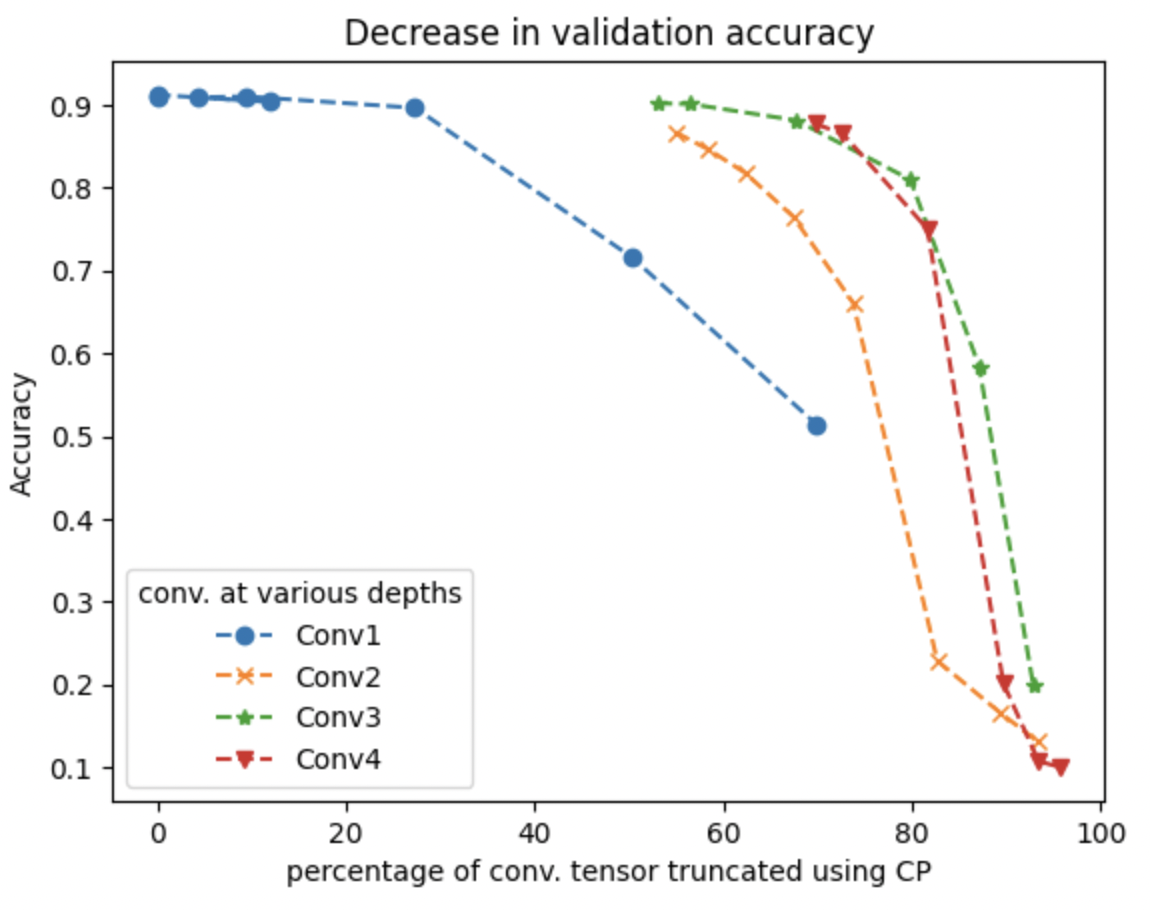}
\includegraphics[width=7cm]{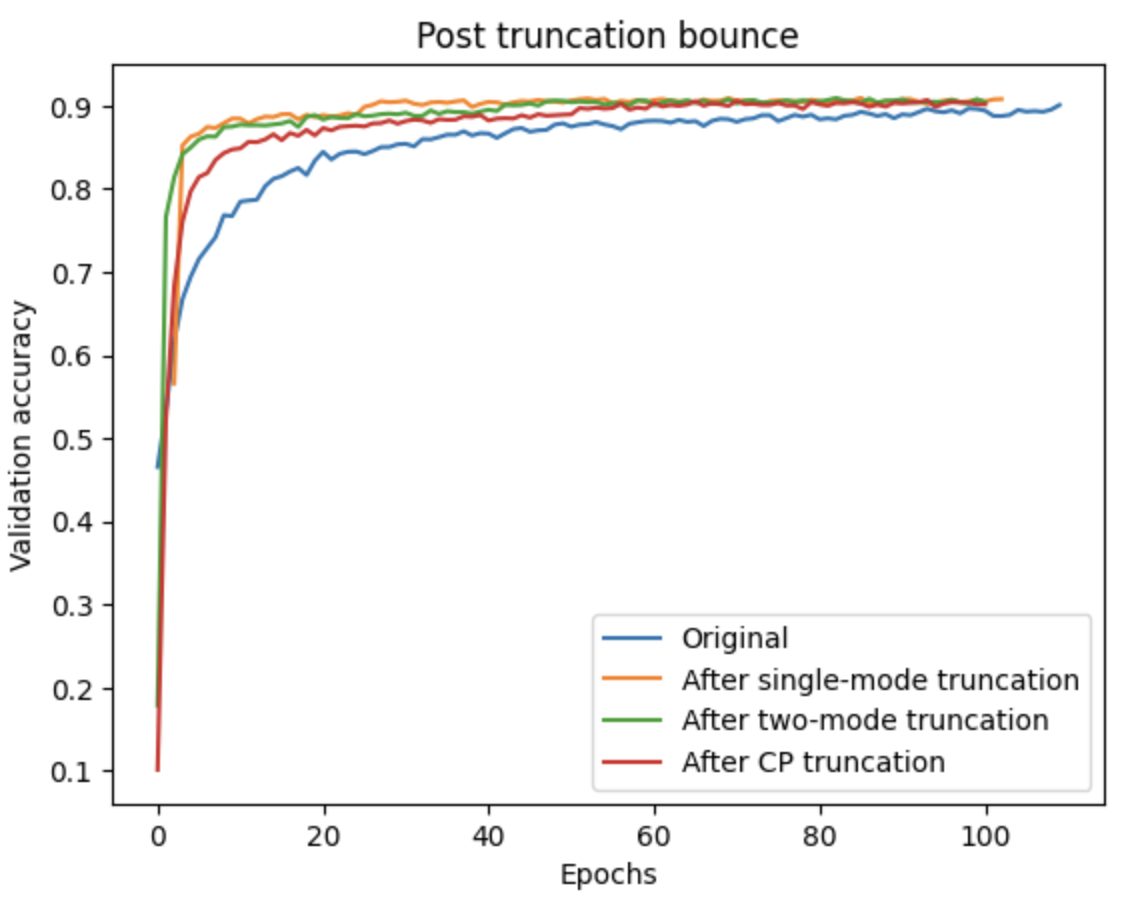}
\caption{(a) Impact of CP truncation on the model's accuracy. The ranges of values for the CP rank for the four layers are $\{120, 100, 80, 60, 40, 20, 10, 5\}, \{120, 100, 80, 60, 40, 20, 10, 5\},$ $\{120, 100, 50, 20, 10, 5\}, \{120, 100, 50, 20, 10, 5\}$  (b) Bounce back in validation accuracy when training the truncated model. We performed an extreme truncation in each of the three scenarios listed in the plot. For instance, for the two-mode truncation case, we truncated correlations across the bipartition between the input and output channels and the remaining indices in all four convolution layers, keeping $\phi = 22, 22, 7$ and $7$, respectively, which decreased the validation accuracy to $0.1787$. The validation accuracy bounced back to $0.7661$ after the first epoch and then to $0.8139, 0.8416, 0.8416, 0.8498, 0.8594$ in the next five epochs. The CP truncation was carried out with rank $r = 10$ on all four convolution layers, which dropped the validation accuracy to $0.1011$. The accuracy returned to $0.5215, 0.6826, 0.7606, 0.7972, 0.8143$ in the first five epochs. However, we found the bounce was slower in this case than in the two cases with correlation truncation.}\label{fig:cp_and_bounce}
\end{figure}

We first applied correlation truncation across single-mode bipartitions OUT, IN, KW, and KH. We replaced the convolution kernels in the model with their truncated versions and thus obtained a truncated model. We then assessed the impact of the truncation on the training and validation accuracy of the model.

The results of single-mode correlation truncation are shown in Fig.~\ref{fig:trunc_single_modes}. We make several observations. The plots show that validation accuracy generally decreases with increased truncation magnitude (i.e., the norm loss, Eq.~\ref{eq:normloss}). We also found similar accuracy trends for the training dataset (not shown here). However, the impact of truncation on accuracy varies with the choice of mode and depth in the CNN. Firstly, truncating correlations across smaller bipartitions, namely, the kernel width and height, resulted in a greater loss of accuracy. This is expected since smaller bipartitions have fewer singular values; therefore, each singular value captures a larger proportion of the kernel norm. In contrast, single-mode truncation of larger bipartitions is relatively less impactful, especially at deeper layers. For instance, we see that the number of in and out channels can be substantially truncated (discarding up to nearly half the kernel norm) without significantly losing accuracy. Fig.~\ref{fig:mode_pair_trunc} shows the results of two-mode correlation truncation. We find again that large truncations are possible at deeper layers without significant loss in accuracy. These results indicate that it is possible to compress the convolution kernels across certain cuts without significant loss of accuracy, even though the learning algorithm was not explicitly biased for such resilience.

The plots shown in Fig.~\ref{fig:depth_scaling} directly compare the impact of truncating a layer across a particular bipartition against the depth of the layer in the network. We find that deeper layers are generally more resilient against truncations.

The top panel of Fig.~\ref{fig:cp_and_bounce} shows the results for CP decomposition-based truncation. We see that for a given norm loss, the CP decomposition generally results in a larger loss of accuracy when compared to SVD-based truncation. This suggests that SVD-based truncation --- which discards minimal correlations for fixed norm reduction --- can achieve a greater lossless compression of convolution kernels compared to CP decomposition.

Finally, we re-trained the truncated models to assess how quickly the truncated model recovers the accuracy of the original model. The results are shown in Fig.~\ref{fig:cp_and_bounce} (bottom panel). We performed aggressive truncations of the model using single-mode correlation, two-mode correlation, and CP-based truncations. In each scenario, we applied the truncation to all four convolution kernels in the model such that the loss in norm was extreme, preserving only a small fraction of the original kernel. We expected these truncated models to be close to randomly initialized ones. Note, however, that the fully connected layers were unaffected by the truncation. Remarkably, in all truncation scenarios, the accuracy bounced back to close to pre-truncation accuracy only after a few epochs of training, with SVD-based truncated models bouncing back faster than the CP truncated ones. This suggests that for this model these truncations do not transport the model to a worse minimum.

\end{appendix}





\bibliography{references.bib}


\end{document}